\documentclass{article}

\usepackage{arxiv}
\usepackage{times}
\usepackage[utf8]{inputenc}
\usepackage[T1]{fontenc}
\usepackage{authblk}
\usepackage{natbib} 
\setcitestyle{authoryear,round,citesep={;},aysep={,},yysep={;}}

\usepackage{amsmath,amsfonts,bm}

\def\eqref#1{equation~\ref{#1}}

\def\1{\bm{1}}

\DeclareMathAlphabet{\mathsfit}{\encodingdefault}{\sfdefault}{m}{sl}
\SetMathAlphabet{\mathsfit}{bold}{\encodingdefault}{\sfdefault}{bx}{n}

\usepackage{amsmath, amssymb, mathtools}
\usepackage{amsfonts}
\usepackage{nicefrac}
\usepackage{microtype}

\usepackage{booktabs}
\usepackage[table]{xcolor}
\usepackage{graphicx}
\usepackage{float}
\usepackage{wrapfig}
\usepackage{multirow}
\usepackage{makecell}
\usepackage{tabularx}
\usepackage{rotating}

\usepackage{algorithm}
\usepackage{algpseudocode}
\usepackage{enumitem}

\usepackage{hyperref}
\usepackage{url}
\usepackage{csquotes}
\usepackage{todonotes}

\newcommand{\norman}{\hyperref[data:Norman19]{\texttt{Norman19}}}
\newcommand{\srivatsan}{\hyperref[data:Srivatsan20]{\texttt{Srivatsan20}}}
\newcommand{\jiang}{\hyperref[data:Jiang24]{\texttt{Jiang24}}}
\newcommand{\transfer}{{\textit{covariate transfer}}}
\newcommand{\combo}{{\textit{combo prediction}}}

\newif\ifnonblind
\nonblindtrue
\title{PRiMeFlow: capturing complex expression heterogeneity in perturbation response modelling}

\begin{document}






\author[1*$\dagger$]{Zichao Yan}
\author[1*$\mathsection$]{Yan Wu} 
\author[1*]{Mica Xu Ji}
\author[1*]{Chaitra Agrahar}
\author[1*]{Esther Wershof}
\author[1*]{Marcel Nassar}
\author[1*]{Mehrshad Sadria}
\author[1*]{Ridvan Eksi}
\author[1]{Vladimir Trifonov}
\author[1]{Ignacio Ibarra}
\author[1$\ddagger$]{Telmo Felgueira} 
\author[1$\|$]{Błażej Osiński}     
\author[1$\dagger$]{Rory Stark}

\affil[1]{Altos Labs}

\date{}

\maketitle

\renewcommand{\thefootnote}{\fnsymbol{footnote}}

\footnotetext[1]{Equal contribution.}
\footnotetext[2]{Corresponding authors: \texttt{\{zyan, rstark\}@altoslabs.com}}
\footnotetext[4]{Current affiliation: Verge Genomics}
\footnotetext[3]{Current affiliation: Loka}
\footnotetext[6]{Current affiliation: Independent Researcher}

\renewcommand{\thefootnote}{\arabic{footnote}}

\begin{abstract}
Predicting the effects of perturbations in-silico on cell state can identify drivers of cell behavior at scale and accelerate drug discovery. However, modeling challenges remain due to the inherent heterogeneity of single cell gene expression and the complex, latent gene dependencies. Here, we present PRiMeFlow, an end-to-end flow matching based approach to directly model the effects of genetic and small molecule perturbations in the gene expression space. The distribution-fitting approach taken by PRiMeFlow enables it to accurately approximate the empirical distribution of single-cell gene expression, which we demonstrate through extensive benchmarking inside PerturBench. Through ablation studies, we also validate important model design choices such as operating in gene expression space and parameterizing the velocity field with a U-Net architecture.
\textcolor{black}{Finally, by scaling PRiMeFlow to a broad perturbation data atlas spanning multiple datasets and employing a carefully designed pretraining-finetuning strategy, we demonstrate its outstanding performance on the H1 human embryonic stem cells from the ARC Virtual Cell Challenge benchmark.}

\end{abstract}


\section{Introduction}

Novel imaging and genomics methods have enabled high-throughput small molecule and genetic perturbation technology with rich imaging and single cell gene expression measurements \citep{Zhang2025,dixit2016perturb,chandrasekaran2023jump}. However despite these advances, the space of potential small molecule and genetic perturbations remains massive, especially since perturbations depend on covariates such as cell type and many key cell state shifts only occur in response to multiple perturbations \citep{watanabe2019combinatorial,wei2009klf4}. 

Machine learning models that can predict counterfactual response to perturbations are thus valuable in unlocking in-silico screens that can prioritize the most promising sets of perturbations for experimental testing. To this end, significant benchmarking efforts have emerged in the literature space, for example PerturBench~\citep{wu2024perturbench}. Its novel rank-based metrics have been widely adopted in the field, for example in the first annual Virtual Cell Challenge (ARC VCC,~\citet{roohani2025virtual}). We provide background information on PerturBench and the Virtual Cell Challenge in Section~\ref{sec:perturbench_background}.


Empirical results from PerturBench show that, surprisingly, simpler model variants could outperform more complex versions. For example, ablating the adversarial loss from CPA \citep{lotfollahi2023predicting} and the sparsity constraint from SAMS-VAE \citep{bereket2023modelling} improve \transfer{} task performance.
This suggests that while strong inductive biases are interesting from a modelling perspective and can be performant in specific test settings, design choices should be rigorously ablated as simpler algorithms can generalize better in practice. 

Motivated by the insights of PerturBench, we seek to design an expressive yet conceptually simple generative model to better capture the heterogeneous cellular responses to perturbations. We present PRiMeFlow~\footnote{PRiMe stands for: \textbf{P}erturbation \textbf{R}esponse [\textbf{i}] \textbf{M}od\textbf{e}ling} which is the first of its kind to apply end-to-end learning principles to flow matching for perturbation response modeling. We intentionally make the design choice of eliminating the need for fixed pretrained latent embeddings, typically derived from either PCA or single-cell foundation models~\citep{klein2025cellflow, palla2025scalable}. While these methods compress expression profiles into more tractable lower-dimensional spaces, they risk losing critical biological signals for accurate downstream prediction, as well as error accumulation in the multi-stage inference pipeline.

Distinctively, PRiMeFlow directly operates in the gene expression space and employs a U-Net architecture to parameterize the velocity field. We formulate our contributions in this work as follows:
\begin{enumerate}[left=0pt,nosep,label=(\arabic*)]
    \item We propose PRiMeFlow, a novel flow matching approach that introduces a straightforward, end-to-end design principle for perturbation response modeling.
    \item We demonstrate PRiMeFlow's superior distribution-fitting capabilities through extensive experiments within PerturBench against a wide array of baselines. Our method achieves significant gains in key distributional metrics, such as the recall of ground truth differentially expressed genes (DEGs), highlighting its ability to accurately model gene expression variance.
    \item Through ablation studies, we validate our core design choices. Specifically, we observe significant gains from three key decisions: parameterizing the velocity field in the gene expression space rather than the latent space, using a U-Net over an MLP, and training with independent coupling instead of optimal transport based coupling. 
    \item \textcolor{black}{We experiment scaling PRiMeFlow to large perturbation data atlas spanning multiple datasets, and through a carefully designed pretraining-finetuning strategy, we demonstrate the outstanding performance of PRiMeFlow on the H1 human embryonic stem cells (hESCs) from the Arc Virtual Cell Challenge~\citep{roohani2025virtual}}
\end{enumerate}

\section{Background}

\subsection{Benchmarking virtual cell models in PerturBench}
\label{sec:perturbench_background}

\begin{table}[!t]
\caption{Summary of benchmarking datasets.}
\begin{center}
\begin{tabular}{ l c c c  r c }
 \multirow{1}{*}{Dataset} & Single / Dual & Modality & Biological & Cells & Task \\
                  & perturbations & & states & & \\
 \toprule
 \srivatsan{} & 188 / 0 & small molecule & 3 & \texttt{178,213} & \transfer{} \\
 \jiang{} & 219 / 0 & genetic & 30 & \texttt{1,628,476} & \transfer{} \\
 \norman{}    & 155 / 131 & genetic & 1 & \texttt{91,168} & \combo{} \\
\end{tabular}
\end{center}
\label{tab:datasets}
\end{table}

To rigorously evaluate the performance of PRiMeFlow, we use the PerturBench platform~\citep{wu2024perturbench} which contains well-defined tasks, numerous perturb-seq datasets and diverse evaluation metrics. PerturBench includes widely-adopted, pseudobulk-level metrics such as cosine similarity of log fold-change (LogFC) and root mean square error (RMSE).
To evaluate the model ability of generating biologically meaningful single-cell populations, it also provides distribution-based metrics such as maximum mean discrepancy (MMD) in gene expression (GEX) or PCA space, and differentially expressed genes (DEG) recall. 
Finally, PerturBench provides rank-based versions of these metrics which quantify predictions specificity. 

PerturBench also includes preprocessed perturb-seq datasets, some of which are selected in this work for benchmarking purposes. Refer to Table~\ref{tab:datasets} and Appendix section~\ref{sec:dataset_information_appendix} for details. In particular, we benchmark our models in \srivatsan{} and \jiang{} datasets under the \transfer{} task, and in \norman{} dataset under the \combo{} task.

\subsection{ARC Virtual Cell Challenge 2025}

\textcolor{black}{In 2025, the ARC institute hosted the first annual Virtual Cell Challenge with a task setup that resembles Perturbench's \transfer{}. 300 perturbations from H1 hESCs are selected with stratified effect sizes, and partitioned into training (150), public test (50), and private test (100) sets~\citep{roohani2025virtual}. The competition permits the use of public or proprietary data~\footnote{\url{https://virtualcellchallenge.org/datasets}}.}

\textcolor{black}{The competition attracted over 1200 teams worldwide and received over 300 final submissions. Initial evaluations relied on mean absolute error (MAE), differential expression score (DES), and a PerturBench-motivated perturbation discrimination score (PDS). Later on, a broader set of evaluation metrics was introduced to establish a separate Generalist leaderboard, where models are ranked in each of the seven metrics in total and the average rank is used a holistic assessment of model generalization capability. Details of the expanded metric set can be found in Appendix section~\ref{sec:app_arc_vcc}.}

\subsection{Flow Matching}

Flow matching belongs to a family of generative models which transport a known source distribution ($p_0$, e.g., simple Gaussian distribution) to the unknown data distribution ($p_{data}$), using a neural network parameterized mapping $\psi^\theta$. The goal is to align the transformed distribution (i.e., \textit{push-forward} $\psi^\theta_{\#}p_0$) to $p_{data}$, traditionally achieved by minimizing the negative data log-likelihood~\citep{rezende2015variational}: $L(\theta) = - \sum_i^{n} \log( \psi^\theta_{\sharp}p_0 (x_i) )$. 

A particular example is the continuous normalizing flow (CNF)~\citep{chen2018neural} which parameterizes the time-dependent velocity field $v^\theta_{t}(x)$ with neural networks. By enforcing Lipschitz continuity constraint in $x$ as well as continuity constraint in $t$, it uniquely defines a diffeomorphic mapping $\psi^\theta_{t}$ through the ordinary differential equation (ODE):
\begin{gather*}
    \frac{d}{dt} \psi^\theta_t (x) = v^\theta_t (\psi^\theta_t (x)),\quad\psi^\theta_0 (x) = x
\end{gather*}
At inference time, the learned velocity is used iteratively to solve the ODE, given initial conditions which are samples from $p_0$, where the resulting ODE integration ideally leads to high quality samples that capture the diversity in $p_{data}$. We denote the intermediate probability densities as $p_t$ 
and consider $v_t$ generating $p_t$ if the two quantities satisfy the continuity equation:
\begin{gather*}
    \partial p_t + \nabla \cdot (p_t v_t) = 0
\end{gather*}
Advantages aside, the training of CNF is impeded by expensive ODE simulation required to compute the log-likelihood. This is where a flow matching approach offers a more efficient and simulation-free alternative. 

In flow matching, we first consider designing the probability path $p_t$ that interpolates between $p_0$ and $p_{data}$. This requires selecting a conditioning variable $z$ (e.g., $z \sim p_{data}$) and the conditional probability path $p_t(x|z)$ that can be used to recover $p_t$ through a weighted aggregation of the conditional paths: $p_t(x) = \int p_t (x|z) p(z) dz$. 

It is important that $p_t (x|z)$ is selected to induce simple closed-form expressions in its corresponding conditional velocities $v_t(x|z)$. In the example of $z \sim p_{data}$ and assuming $p_0$ is the standard Gaussian distribution, $p_t (x|z)$ takes the form of $\mathcal{N}(x|t\cdot z,(1-t)^2\cdot I)$, and through the continuity equation, it can be shown that $v_t(x|z)=\frac{z-x}{1-t}$.

A further result from~\cite{lipman2022flow} shows that the ground truth generating velocity of $p_t$, referred to as the marginal velocity $v_t(x)$, is connected to $v_t(x|z)$ in the following way:
\begin{align*}
    v_t(x) = \int v_t(x|z) \frac{p_t(x|z)p(z)}{p_t(x)} dz
\end{align*}
It is obvious that the marginal velocity is intractable, hence the difficulty of directly using it as a regression target for the learned velocity field, as in the naive flow matching loss:
\begin{gather}
    \mathcal{L}_{FM} = \mathbb{E}_{t, x\sim p_t}||v_t^\theta(x) - v_t(x)||^2 \label{eq:fm}
\end{gather}
And this leads to the conditional flow matching loss proposed in~\cite{lipman2022flow}:
\begin{gather}
    \mathcal{L}_{CFM} = \mathbb{E}_{t, z\sim p_z, x\sim p_t(x|z)}||v_t^\theta(x) - v_t(x|z)||^2 \label{eq:cfm}
\end{gather}
which is a tractable alternative to $\mathcal{L}_{FM}$. It has been shown in~\cite{lipman2022flow} Theorem 2 that $\nabla_\theta \mathcal{L}_{CFM} =\nabla_\theta \mathcal{L}_{FM}$, thus optimizing the tractable Eq.~\ref{eq:cfm} leads to the same gradient as Eq.~\ref{eq:cfm}, making both effectively equivalent in expectation. The conditional flow matching loss has been widely applied in generative modeling contexts and has offered a wide selection of conditioning variables and conditional probability paths, such as nonlinear interpolations, couplings between empirical data distributions, and optimal transport~\citep{albergo2023stochastic,liu2022flow,tong2023improving}.

\begin{figure}[!t]
\centering
\includegraphics[width=1.\textwidth]{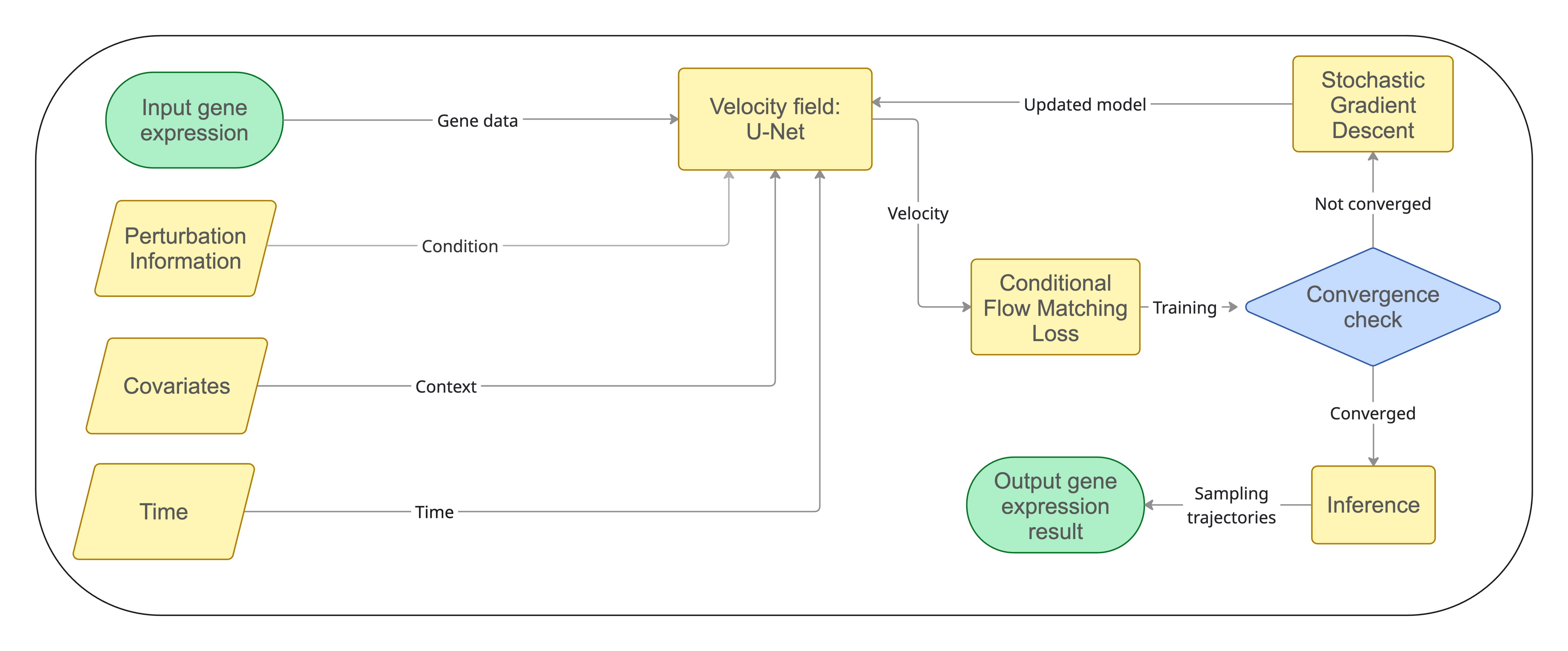}
\caption{PRiMeFlow workflow overview. The velocity field is directly parameterized in the gene expression space, using a U-Net architecture, and is conditioned on information of the perturbation, covariates, as well as the time variable. The velocity field is learned with the standard conditional flow matching loss. At the inference time, perturbed cells are sampled by solving the initial value problems using the learned velocity field.}
\label{fig:workflow}
\end{figure}

\section{Method}

For a perturb-seq dataset $\mathcal{D}$, consider the set of $\mathrm{log1p}$ normalized gene expression profiles $\mathcal{X}=\{x_i \in \mathbb{R}^m \}_{i=1}^n$ where $n$ denotes the size of the dataset and $m$ denotes the number of expression genes. The identity of the genes, for example gene names, is denoted by $\mathcal{G}=\{g_j\}_{j=1}^m$. 

Every expression profile corresponds to a cell with condition $c$ specifying genetic/small-molecule perturbations $p$, and covariates $cov$ such as cell subtype and other cell treatments which are dataset specific. We denote the set of conditions with $\mathcal{C}=\{c_i\}_{i=1}^n$ where $c_i=(\{p_{i, k}\}_{k=1}^{n_{i}}, cov_i)$. Given the possibility of combinatorial perturbations, each condition may contain a plurality of perturbations as indicated by the set $\{p_{i, k}\}_{k=1}^{n_{i}}$ where $n_i$ denotes the number of perturbations. An empty perturbation set ($n_i=0$) indicates a control cell.

\subsection{Direct flow matching in gene expression space}

In this section, we introduce PRiMeFlow, our perturbation response model. A simplified view of the PRiMeFlow workflow is shown in Figure~\ref{fig:workflow}. The design philosophy behind PRiMeFlow is straightforward: to tackle the heterogeneity of gene expression measurements (both inter- and intra-condition) and the complex interaction relationships among genes, we propose to directly parameterize the velocity field $v_t^\theta$ in the gene expression space.

While this exposes the velocity field to the full complexity in the raw gene expression data, it also creates an opportunity for the model to learn dynamics that directly govern the transition of expression profiles, especially when the velocity field is parameterized with an architecture that is adequate for capturing the noise and intricate structures inherent in gene expression data.

In particular, $v_t^\theta$ takes as input three main components:
\begin{itemize}[leftmargin=*,itemsep=2pt, parsep=0pt]
    \item Gene expression profile $x_t$ and gene identities $\mathcal{G}$. $x_t$ is sinusoidally encoded to add a feature dimension i.e., $\mathrm{Enc}(x_t) \in \mathbb{R}^{m \times d}$. $\mathcal{G}$ is included to ingest biological interpretation in the first dimension in $x_t$ and is mapped to learnable embedding vectors: $\mathrm{Emb}(\mathcal{G}) \in \mathbb{R}^{m \times d}$;
    \item Condition $c_i=(\{p_{i,k}\}_{k=1}^{n_{i}}, cov_i)$. We use a compositional encoding of potentially combinatorial perturbations and one-hot encoding for covariate: $\mathrm{Enc}(c_i)=[\sum_{k}\mathrm{OneHot}(p_{i,k}) || \mathrm{OneHot}(cov_{i})]$, where $[...||...]$ denotes concatenation along the feature dimension;
    \item Time $t$ which is also sinusoidally encoded in $\mathrm{Enc}(t)$.
\end{itemize}
The output of the velocity field is in the following form:
\begin{align}
    \hat{v}=v_t^\theta(\mathrm{Enc(x_t)}+\mathrm{Emb}(\mathcal{G}), \mathrm{Enc(c_i)}, \mathrm{Enc}(t))\label{eq:model_line}
\end{align}
$v_t^\theta$ is parameterized with a U-Net architecture which we have found crucial for its performance. We have implemented our U-Net based on OpenAI's guided diffusion library~\citep{dhariwal2021diffusion}.

\newtheorem{remark}{Remark}
\begin{remark}[Architecture choice for velocity field]
   U-Nets are designed to capture spatial relationships, such as the relationship between spatial coordinates in imaging data, which do not apply to gene expression vectors where the ordering of genes is arbitrary. While theoretically suboptimal compared to permutation-invariant architectures like Transformer~\citep{vaswani2017attention}, we found the U-Net to be a pragmatic choice: it is highly parameter efficient compared to the Transformer, requires significantly lower computational resources, and empirically outperforms MLP.
\end{remark}

\begin{wrapfigure}{r}{0.5\columnwidth} 
\vspace{-2em} 
\begin{minipage}{\linewidth}
\begin{algorithm}[H] 
\caption{PRiMeFlow Training}
\label{alg:primeflow_training}
\begin{algorithmic}
\Require $p_0$, $p_{data}$, $\theta$, $\mathcal{G}$, $p_{uncond}$, noise schedule $\sigma$, batch size $b$, learning rate $\eta$.
\State Initialize parameters $\theta$.
\While{not converged}
\State Sample $x_1 \in \mathbb{R}^{b\times m} \sim p_{data}$.
\State Retrieve conditions $C=\{c_i\}_{i=1}^{b}$, \\\quad\quad\quad\,\, where $c_i \leftarrow \emptyset$ with prob $p_{uncond}$. 
\State Sample $x_0 \in \mathbb{R}^{b\times m} \sim p_0$.
\State Sample time steps $t \in \mathbb{R}^{b} \sim \text{Uniform}(0, 1)$.
\State Sample  $x_t \in \mathcal{N}(x\,|\,(1-t)x_0 + tx_1 , \sigma^2 \mathcal{I})$.
\State Compute target $v_t(x|(x_0,x_1)) = x_1 - x_0$.
\State Predict velocity $\hat{v}$ as Eq.~\ref{eq:model_line}.
\State Compute $\mathcal{L}_{CFM}$ as Eq.~\ref{eq:cfm}.
\State Update parameters: $\theta \leftarrow \theta - \eta \nabla_\theta \mathcal{L}_{CFM}(\theta)$.
\EndWhile
\end{algorithmic}
\end{algorithm}
\end{minipage}
\vspace{-10pt} 
\end{wrapfigure}


\subsection{Training PRiMeFlow and ablations}

To train PRiMeFlow, we construct probability paths as independent couplings between $p_0$ and $p_{data}$. We choose $p_0$ to be the standard Gaussian distribution, and $p_{data}$ to be the empirical cell populations that includes both perturbed and control cells. For training step details, refer to Algorithm~\ref{alg:primeflow_training}.

To understand the importance of architecture choices we perform ablation studies. We provide an ablation model referred to as PRiMeFlow MLP which parameterizes the velocity field with an MLP instead of U-Net. The training steps of PRiMeFlow MLP remain unchanged.

To demonstrate the value of end-to-end learning, we provide an alternative implementation of PRiMeFlow that operates in the PCA space of gene expression, with and without optimal transport. These baselines are referred to as FlowMatching PCA and FlowMatching PCA OT respectively. While FlowMatching PCA still uses a standard Gaussian as $p_0$, FlowMatching PCA OT uses empirical control cells as $p_0$ and is trained with optimal transport-based data-coupling. Both FlowMatching PCA and FlowMatching PCA OT use MLP for the velocity field. The training steps of FlowMatching PCA OT are altered to take data coupling into account and, in effect, become similar to CellFlow~\citep{klein2025cellflow}.

\section{Experiments}

\begin{table}[ht]
\caption{Distributional metric performance of PRiMeFlow on the \srivatsan{} dataset.}
\centering
\begin{tabular}{@{}lccccc@{}}
\toprule
\multirow{2}{*}{Model} & MMD & MMD & MMD & DEG\\
                       & GEX & GEX rank & PCA & recall \\
 \toprule
PRiMeFlow CFG1 & $\mathbf{0.13 \pm 4 \times 10^{-3}}$ & $\mathbf{0.13 \pm 8 \times 10^{-3}}$ & $\mathbf{0.14 \pm 8 \times 10^{-3}}$ & $\mathbf{0.26 \pm 1 \times 10^{-2}}$ \\
PRiMeFlow CFG3 & $0.25 \pm 2 \times 10^{-3}$ & $0.22 \pm 7 \times 10^{-3}$ & $0.31 \pm 1 \times 10^{-2}$ & $\mathbf{0.27 \pm 6 \times 10^{-3}}$ \\
PRiMeFlow CFG5 & $0.47 \pm 1 \times 10^{-2}$ & $0.29 \pm 2 \times 10^{-2}$ & $0.60 \pm 2 \times 10^{-2}$ & $0.25 \pm 5 \times 10^{-3}$ \\
\midrule
FM PCA CFG1 & $1.2 \pm 7 \times 10^{-3}$ & $0.26 \pm 6 \times 10^{-3}$ & $0.50 \pm 2 \times 10^{-2}$ & $0.042 \pm 2 \times 10^{-3}$ \\
FM PCA CFG3 & $1.3 \pm 5 \times 10^{-3}$ & $0.25 \pm 7 \times 10^{-3}$ & $0.44 \pm 1 \times 10^{-2}$ & $0.071 \pm 3 \times 10^{-3}$ \\
FM PCA CFG5 & $1.4 \pm 4 \times 10^{-3}$ & $0.29 \pm 4 \times 10^{-3}$ & $0.53 \pm 1 \times 10^{-2}$ & $0.10 \pm 3 \times 10^{-3}$ \\
\midrule
CPA & $2.4 \pm 1 \times 10^{-2}$ & $0.30 \pm 9 \times 10^{-3}$ & $0.53 \pm 4 \times 10^{-3}$ & $0.0073 \pm 2 \times 10^{-3}$ \\
CPA (noAdv) & $2.3 \pm 3 \times 10^{-2}$ & $0.25 \pm 5 \times 10^{-3}$ & $0.49 \pm 1 \times 10^{-2}$ & $0.0040 \pm 2 \times 10^{-3}$ \\
SAMS-VAE & $2.5 \pm 2 \times 10^{-2}$ & $0.30 \pm 8 \times 10^{-3}$ & $0.69 \pm 1 \times 10^{-2}$ & $0.00016 \pm 1 \times 10^{-4}$\\
SAMS-VAE (S) & $2.9 \pm 1 \times 10^{-2}$ & $0.28 \pm 5 \times 10^{-3}$ & $0.79 \pm 1 \times 10^{-2}$ & $0 \pm 0$ \\
Biolord & $4.9 \pm 3 \times 10^{0}$ & $0.36 \pm 2 \times 10^{-1}$ & $4.3 \pm 4 \times 10^{0}$ & $4.7e-05 \pm 1 \times 10^{-4}$ \\
\midrule
Latent & $4.3 \pm 2 \times 10^{-1}$ & $0.26 \pm 6 \times 10^{-2}$ & $2.0 \pm 2 \times 10^{-1}$ & $0.0 \pm 0.$ \\
Decoder & $4.2 \pm 4 \times 10^{-3}$ & $0.16 \pm 2 \times 10^{-2}$ & $1.9 \pm 5 \times 10^{-3}$ & N/A \\
Linear & $2.2 \pm 7 \times 10^{-3}$ & $0.31 \pm 1 \times 10^{-3}$ & $0.76 \pm 9 \times 10^{-4}$ & $0.0036 \pm 3 \times 10^{-4}$ \\
\bottomrule
\end{tabular}
\label{tab:srivatsan}
\end{table}


\textcolor{black}{In sections~\ref{sec:perturbench_cov_transfer},~\ref{sec:perturbench_combo} and~\ref{sec:perturbench_figures}, we evaluate PRiMeFlow in PerturBench benchmarks.} Comparison to important prior work is made, such as to CPA, SAMS-VAE, Biolord and their variants which are implemented in PerturBench~\citep{wu2024perturbench}. For simple baselines, we include Latent Additive, Linear Additive, and Decoder-Only models. We also compare to various ablation models such as PRiMeFlow MLP, FM PCA and FM PCA OT.

All models were trained using five random seeds, and we report the mean and standard deviation. Hyperparameters for PRiMeFlow MLP and FM PCA were selected separately through 60-trial HPO sweeps. FM PCA OT reuses FM PCA hyperparameters while adding OT-specific hyperparameters. PRiMeFlow hyperparameters were manually determined due to the model's higher computational cost and our resource constraints. We report the hyperparameters in the Appendix section~\ref{sec:hp}.

To compute the evaluation metrics, we sampled 1,000 single-cell gene expression profiles from each model. For inference in the four flow matching based methods, we used 100 Euler integration steps and explored CFG settings from 1 to 5. All results are reported on the test split. 

\textcolor{black}{In section~\ref{sec:arc_vcc_results}, we evaluate PRiMeFlow on H1 hESCs from the ARC VCC. We employ a carefully designed pretraining-finetuning strategy, first scaling PRiMeFlow across a large perturbation data atlas (VCC+public+proprietary data), then followed by targeted finetuning using a subset of the data to facilitate transfer learning of perturbation effects from external datasets (i.e., any non-VCC data) to public/private test perturbations in H1 hESCs. Due to space constraints, detailed description of our data composition and splits, training and inference strategies, are deferred to Appendix section~\ref{sec:app_arc_vcc}.}

\subsection{Performance of PRiMeFlow in \transfer{} task}
\label{sec:perturbench_cov_transfer}

In this section, we evaluate PRiMeFlow in \srivatsan{} and \jiang{} datasets under the \transfer{} task. The task measures the model's ability to capture the effects of small-molecule and genetic perturbations in novel covariate conditions where those effects are unmeasured in the training data. Results for \srivatsan{} are reported in Table~\ref{tab:srivatsan} which shows a selected set of distributional metrics. Results for the complete distributional metrics, more ablation models such as PRiMeFlow MLP and FM PCA OT, are reported in Table~\ref{tab:srivatsan_full_dist} in the Appendix due to space constraints. The complete results for \jiang{} dataset is reported in Appendix Table~\ref{tab:jiang_full_dist}.

Overall, we observe that PRiMeFlow (CFG=1) achieves state-of-the-art performance in both datasets across major distributional metrics which are MMD in either GEX or PCA space, DEG recall as well as DEG recall rank. In \srivatsan{} dataset, PRiMeFlow achieves SOTA performance in MMD GEX rank while maintaining a competitive edge in MMD PCA rank. The converse is true in \jiang{}. Increasing the CFG weight from 1 to 5 generally degrades distributional fit. However, stronger guidance appears to yield marginal improvements in DEG recall and its rank metrics in \srivatsan{} dataset. This holds true for both PRiMeFlow and FM PCA, but not for PRiMeFlow MLP or FM PCA OT.

In particular, comparing PRiMeFlow to FM PCA in both datasets suggests that operating end-to-end in the full gene expression space and directly updating the gene expression vectors through time with ODE integration results in a significantly closer fit to the ground truth distribution of gene expression. An immediate benefit of this approach is the explicit modeling of gene expression variance, which translates directly to a higher recall of ground-truth DEGs. Furthermore, comparing FM PCA to FM PCA OT shows that optimal transport based data coupling did not meaningfully add to model performance.

We compare PRiMeFlow to PRiMeFlow MLP in order to understand the effect of architecture choices. Overall, the MLP variant struggles to achieve decent fits on distributional or traditional pseudobulk based metrics. This result suggests that the U-Net architecture possesses an advantageous inductive bias compared to MLP for modelling complex latent dependencies in raw gene expression data.

Finally, we report traditional metrics, such as LogFC, RMSE and their rank versions, for \srivatsan{} and \jiang{} datasets in Appendix Table~\ref{tab:srivatsan_additional_pseudo} and Table~\ref{tab:jiang_additional_pseudo}. While PRiMeFlow remains generally competitive, it is outperformed by earlier works such as CPA and SAMS-VAE on those pseudobulk-based metrics. This highlights a crucial trade-off: while earlier models excel at predicting mean expression, PRiMeFlow's unique advantage lies in capturing the higher-order characteristics of the ground-truth gene expression distribution. This distinction also illustrates how metric selection fundamentally shapes performance evaluation and the definition of model success.


\subsection{Performance of PRiMeFlow in \combo{} task}
\label{sec:perturbench_combo}

For the \combo{} task, we use the \norman{} dataset and report the results in Table~\ref{tab:norman}. Pseudobulk-based metrics are reported in Appendix Table~\ref{tab:norman_additional_pseudo}. Full distributional metrics and additional ablation models such as PRiMeFlow MLP and FM PCA OT are shown in Appendix Table~\ref{tab:norman_full_dist}. 

Similar to our previous observations, PRiMeFlow demonstrates a significantly higher distribution-fitting capability and remains competitive on MMD GEX/PCA rank metrics amongst all baselines, despite being outperformed by simpler baselines such as Decoder-Only and Latent Additive models on a subset of metrics. This highlights the ongoing need for improved combinatorial prediction capabilities in perturbation response modeling.

Through extensive comparison to ablated models in Appendix Table~\ref{tab:norman_full_dist}, we are also able to make similar conclusions regarding the benefits of leveraging U-Net architectures, operating in full gene expression space, as well as simple independent coupling between standard Gaussian distribution $p_0$ and $p_{data}$.

Overall, it is important to point out that throughout Table~\ref{tab:srivatsan} and~\ref{tab:norman}, PRiMeFlow is the only method to achieve a decent recall performance on the top 50 ground truth DEGs, which is explained by its superior capability for capturing the full gene expression distribution on a single-cell level.

\begin{table}
\caption{Distributional metric performance of PRiMeFlow on the \norman{} dataset.}
\centering
\begin{tabular}{@{}lcccccc@{}}
\toprule
\multirow{2}{*}{Model} & MMD & MMD & MMD & DEG\\
                       & GEX & GEX rank & PCA & recall \\
 \toprule
PRiMeFlow CFG1 & $\mathbf{0.26 \pm 7 \times 10^{-3}}$ & $0.019 \pm 4 \times 10^{-3}$ & $\mathbf{0.34 \pm 2 \times 10^{-2}}$ & $0.54 \pm 2 \times 10^{-2}$ \\
PRiMeFlow CFG3 & $0.61 \pm 3 \times 10^{-2}$ & $0.066 \pm 9 \times 10^{-3}$ & $0.86 \pm 5 \times 10^{-2}$ & $\mathbf{0.61 \pm 8 \times 10^{-3}}$ \\
PRiMeFlow CFG5 & $1.3 \pm 4 \times 10^{-2}$ & $0.17 \pm 1 \times 10^{-2}$ & $1.8 \pm 5 \times 10^{-2}$ & $0.56 \pm 1 \times 10^{-2}$ \\
\midrule
FM PCA CFG1 & $0.77 \pm 4 \times 10^{-3}$ & $0.059 \pm 3 \times 10^{-3}$ & $0.61 \pm 5 \times 10^{-3}$ & $0.37 \pm 2 \times 10^{-3}$ \\
FM PCA CFG3 & $1.1 \pm 8 \times 10^{-3}$ & $0.098 \pm 2 \times 10^{-3}$ & $0.82 \pm 9 \times 10^{-3}$ & $0.45 \pm 2 \times 10^{-3}$ \\
FM PCA CFG5 & $1.9 \pm 4 \times 10^{-2}$ & $0.18 \pm 1 \times 10^{-2}$ & $1.6 \pm 4 \times 10^{-2}$ & $0.43 \pm 4 \times 10^{-3}$ \\
\midrule
CPA & $5.6 \pm 3 \times 10^{-2}$ & $0.14 \pm 9 \times 10^{-3}$ & $2.2 \pm 2 \times 10^{-2}$ & $0.032 \pm 4 \times 10^{-3}$ \\
CPA (noAdv) & $5.5 \pm 1 \times 10^{-1}$ & $0.13 \pm 2 \times 10^{-2}$ & $2.2 \pm 1 \times 10^{-1}$ & $0.016 \pm 9 \times 10^{-3}$ \\
SAMS-VAE & $4.1 \pm 4 \times 10^{-2}$ & $0.034 \pm 1 \times 10^{-2}$ & $1.9 \pm 3 \times 10^{-2}$ & $0 \pm 0$ \\
SAMS-VAE (S) & $3.3 \pm 5 \times 10^{-2}$ & $0.088 \pm 6 \times 10^{-3}$ & $0.74 \pm 5 \times 10^{-2}$ & $0.028 \pm 6 \times 10^{-3}$ \\
Biolord & $2.8 \pm 5 \times 10^{-3}$ & $0.066 \pm 3 \times 10^{-3}$ & $1.6 \pm 5 \times 10^{-3}$ & $0 \pm 0$ \\
\midrule
Latent & $6.7 \pm 4 \times 10^{-3}$ & $0.012 \pm 1 \times 10^{-3}$ & $3.2 \pm 6 \times 10^{-3}$ & $0.00043 \pm 3 \times 10^{-4}$ \\
Decoder & $6.7 \pm 6 \times 10^{-3}$ & $\mathbf{0.0090 \pm 8 \times 10^{-4}}$ & $3.2 \pm 4 \times 10^{-3}$ & N/A \\
Linear & $2.5 \pm 4 \times 10^{-2}$ & $0.026 \pm 2 \times 10^{-3}$ & $1.2 \pm 4 \times 10^{-2}$ & $0.018 \pm 2 \times 10^{-3}$ \\
\bottomrule
\end{tabular}
\label{tab:norman}
\end{table}

\begin{figure}[!t]
\centering
\includegraphics[width=1.\textwidth]{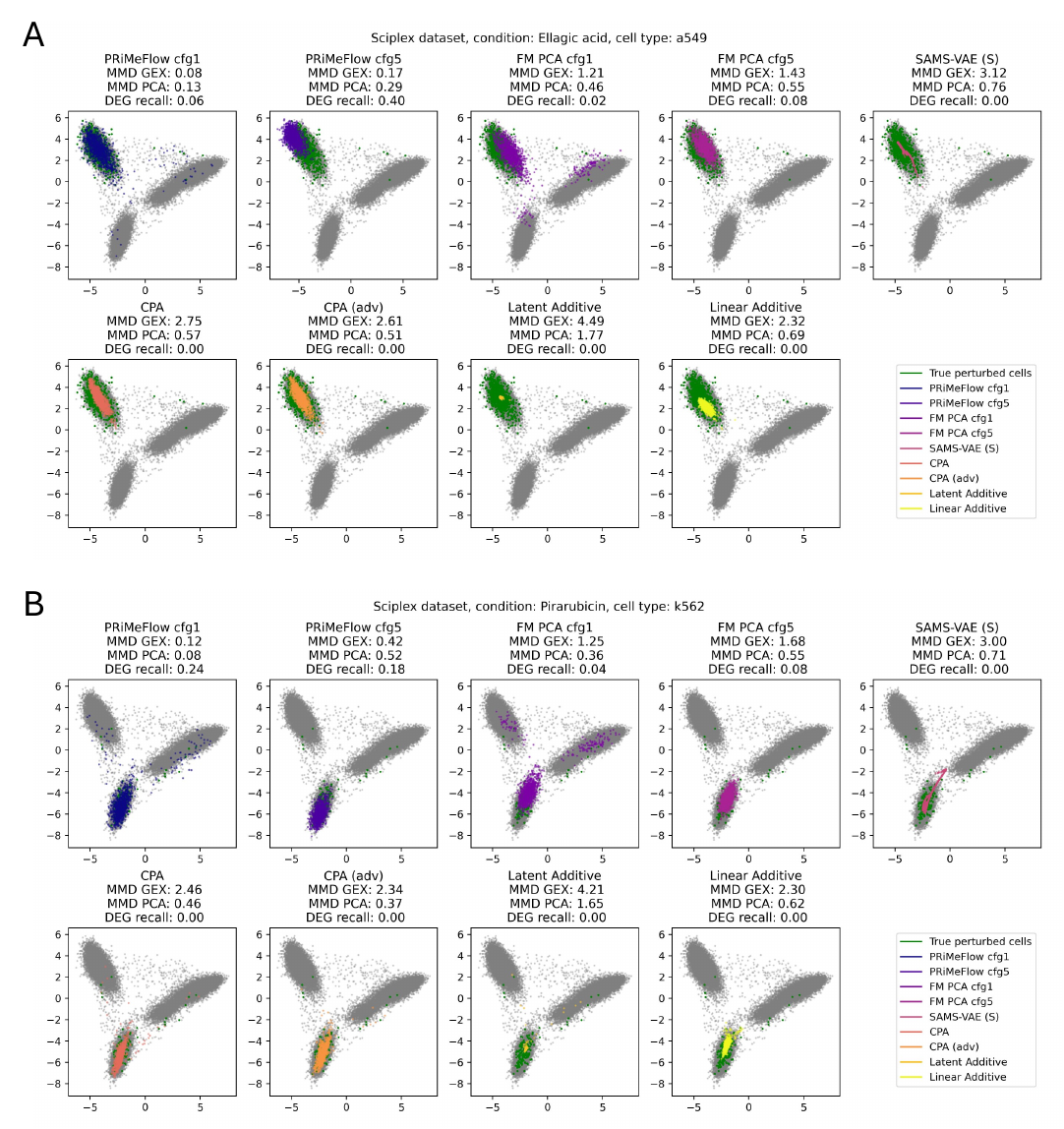}
\caption{Visualization of cells (ground truth and model predictions) in \srivatsan{} test split. Gray marks all the cells in the test split, used for computing PCA and illustrated as background. Green cells are the ground truth perturbed cells for one condition, repeated in each subplot. From each model, 100 cells are sampled, then projected to the same PCA space. \textsf{\bfseries A}): Ellagic acid perturbation in a549 cell line; \textsf{\bfseries B}) Pirarubicin perturbation in k562 cell line.}
\label{fig:srivatsan_pca_scatter}
\end{figure}

\subsection{PRiMeFlow captures heterogeneous gene expressions}
\label{sec:perturbench_figures}

In Figure~\ref{fig:srivatsan_pca_scatter}, we take 1,000 sampled cells from each model and project them into the same PCA space as the reference test data in the \srivatsan{} dataset. The reference cells including control and perturbed cells, cluster naturally into three cell lines that are K562, A549, and MCF-7. We select two random conditions, which are Ellagic acid in a549 and Pirarubicin in k562. The ground truth perturbed cells are colored in green.

In both cases, the projection illustrates how gene expression from PRiMeFlow (CFG=1) is fitted more closely to ground truth perturbed cells at a distributional level compared to baseline methods, confirmed by its superior MMD metrics in both gene expression and PCA space. This suggests PRiMeFlow is less prone to mode collapse, especially compared to traditional methods such as SAMS-VAE and Latent Additive, which struggled to properly cover the whole target distribution. This is in line with observations made in PerturBench~\citep{wu2024perturbench}.

FM PCA also demonstrates strong distribution-fitting capability, but it is worth pointing out that in terms of MMD computed in the PCA space, PRiMeFlow still outperforms FM PCA. Notably, while the visualizations rely on only the top two principal components (PCs), the PCA-based MMD is robustly computed using 30 PCs.

Interestingly, in the case of Ellagic acid in a549, increasing CFG weight from 1 to 5 led to improved DEG recall performance at the cost of other distributional metrics, resulting in a distortion in the visualization. Nevertheless, amplifying the perturbational effect can benefit certain downstream tasks that are focused on perturbation-specific DEGs. Overall, this crucial trade-off highlights the importance of looking at a comprehensive set of evaluation metrics, and that selection of models should be based on needs of the biological tasks. Additional visualization results in \srivatsan{} and \jiang{} datasets are shown in Appendix section~\ref{sec:add_pca_sri} and~\ref{sec:add_pca_jiang}.

\begin{table}[h]
\caption{\textcolor{black}{PRiMeFlow performance on H1 hESCs private test set using all seven ARC VCC metrics.}}
\centering
\resizebox{\textwidth}{!}{
\begin{tabular}{@{}lccccccc@{}}
\toprule
\multirow{1}{*}{Model} & DES $\uparrow$ & PDS $\uparrow$ & MAE $\downarrow$ & SPEARMAN $\uparrow$ & SPEARMAN\_LFC $\uparrow$ & AUPRC $\uparrow$ & PEARSON $\uparrow$ \\
\toprule
PRiMeFlow pretrained & 0.210 & 0.695 & 0.074 & 0.406 & 0.320 & 0.264 & 0.181 \\
PRiMeFlow 300pts & 0.223 & 0.725 & 0.084 & 0.431 & 0.381 & 0.264 & 0.193\\
PRiMeFlow 250pts & 0.228 & 0.747 & 0.086 & 0.473 & 0.396 & 0.266 & 0.203 \\
\midrule
xBio & 0.305 & 0.811 & 0.770 & 0.564 & 0.087 & 0.252 & 0.217 \\
Mean Predictors & 0.305 & 0.741 & 6.723 & 0.294 & 0.213 & 0.582 & 0.217 \\
Shippers & 0.354 & 0.699 & 0.231 & 0.000 & 0.227 & 0.576 & 0.123 \\
\bottomrule
\end{tabular}}
\label{tab:arc_vcc_results}
\end{table}

\subsection{Performance of PRiMeFlow on H1 hESCs from ARC Virtual Cell Challenge}
\label{sec:arc_vcc_results}


\textcolor{black}{In Table~\ref{tab:arc_vcc_results}, we report the performance of PRiMeFlow on the 100 held-out private test perturbations in H1 hESCs. We evaluate our models across the expanded ARC VCC metrics set, and select the top-three performing models from the Generalist leaderboard~\footnote{\url{https://virtualcellchallenge.org/leaderboard}} for comparison. Additional results on the 50 held-out public test perturbations can be found in Appendix section~\ref{sec:vcc_public_test_ret}}.

\textcolor{black}{While the pretrained-only PRiMeFlow demonstrates decent predictive capability, it is significantly outperformed by our two finetuned models: PRiMeFlow-300pts and PRiMeFlow-250pts. For PRiMeFlow-300pts, finetuning takes place on a subset of pretraining data atlas that includes the entire 300 VCC perturbations in H1 hESCs and other external datasets to facilitate transfer learning of the perturbation effects across cell lines and datasets. Similarly, finetuning PRiMeFlow-250pts is further restricted to the 150 training perturbations and the final 100 private test perturbations.}

\textcolor{black}{These results highlight the need of adapting virtual cell models, usually pretrained on large data atlas, to specific cellular contexts and perturbations in order to unlock their full predictive power. Notably, thanks to the \transfer{} task setup, this adaptation can be achieved without additional ground-truth data. Our findings also show that narrowing the finetuning scope to focus more on specific inference-time perturbations will lead to further performance gain.}

\textcolor{black}{While specific baselines can achieve higher scores on individual metrics, our best finetuned model — PRiMeFlow-250pts, stands out by having the lowest average rank overall in the complete Generalist leaderboard. In particular, PRiMeFlow has been shown to achieve strong performance across all metrics while having the minimal distortion from the ground truth cell population, as evidenced by its lower MAE.}

\section{Related Work}


Early approaches such as CPA and SAMS-VAE, use variational autoencoders (VAEs) to learn disentangled latent spaces where basal cellular states and perturbation effects are modeled as independent and compositional components.
CPA \citep{lotfollahi2023predicting} established this framework by decomposing cellular responses into a collection of independent embeddings for basal states, perturbations and covariates. 
A recent follow-up work that is SAMS-VAE~\citep{bereket2023modelling}, introduces sparsity constraints to the latent encodings of perturbation effects and shows improved biological interpretability. Biolord~\citep{biolord} leverages biological attributes such as perturbation and covariates to partition the latent space which leads to biologically meaningful latent representations.

Single-cell foundation models such as scGPT and STATE have leveraged the Transformer architecture and massive pre-training to capture complex gene-gene interactions across millions of cells. scGPT~\citep{scGPT} treats genes as tokens and utilizes a masked value modeling objective to capture complex gene-gene interactions and reconstruct perturbed cellular states. The State~\citep{Adduri2025.06.26.661135} model employs 
a state embedding module that maps 270 million cells into a shared latent space, and a state transition transformer that explicitly models sets of cells and transitions between control and perturbed cell states.

More recent methods have leveraged flow matching and diffusion frameworks to capture the heterogeneity and high-dimensional geometry of cellular responses~\citep{klein2025cellflow,Scvaeder,Squidiff,palla2025scalable,xcell}. CellFlow \citep{klein2025cellflow} utilizes a conditional flow-matching objective to learn time-dependent vector fields, employing an optimal transport based data coupling to encourage parsimonious, shortest-path dynamics between source and target populations. Similarly, scLDM \citep{palla2025scalable} leverages a latent diffusion objective within a compressed token space, utilizing a multi-head cross-attention block (MCAB) to ensure gene permutation invariance and classifier-free guidance to steer the denoising process toward specific attributes. SquiDiff \citep{Squidiff} further refines this generative paradigm by tailoring the diffusion process to the specific manifold of single-cell data, using a score-matching objective to account for the inherent sparsity and technical dropout characteristic of scRNA-seq distributions.

To combat the inherent noise and sparsity of transcriptomic data, frameworks like PRESAGE~\citep{Littman2025.06.03.657653} integrate prior biological knowledge by aggregating gene embeddings from external sources (e.g., gene embeddings derived from protein-protein interaction networks and existing Perturb-seq datasets). 
Another approach which captures the population-level variability using distributional shifts is PerturbDiff~\citep{perturbdiff}, where the entire perturbed population is treated as a random variable, allowing for an accounting of batch effects and other relevant factors. Finally, PerturbGen~\citep{perturbgen} is a trajectory-aware model which aims to understand the trajectory of evolution of the perturbation effects rather than just static effects before and after the perturbation is applied, using a generative architecture to model state-to-state transitions.

\section{Discussion}

PRiMeFlow takes a distinctly different approach from prior models like CellFlow and scLDM. While compressing data into a latent space helps address the high dimensionality and inherent sparsity of single-cell gene expression data, it risks distorting the information in gene expression profiles and discarding crucial biological signals needed to capture subtle, perturbation-specific effects. Instead, PRiMeFlow applies the flow matching framework directly in the full gene expression space, is trained end-to-end, and employs a dedicated U-Net architecture to model the complex underlying gene dependencies. Through extensive benchmarking and ablation experiments, we demonstrate the practical benefits of our new formulation, including a closer fit to the target distribution of perturbed cells, capturing more perturbation-specific effects, and a higher recall of ground truth DEGs.
\ifnonblind
\textcolor{black}{On the H1 hESCs from the ARC Virtual Cell Challenge, these attributes helped our model to achieve outstanding performance across the expanded metric suite.}
\fi

While we have demonstrated the empirical success of the U-Net architecture for velocity field parameterization, exactly why it performs so well remains an open question. U-Net assumes inherent inductive bias of learning sequence order in each input, whereas for gene expression data, the order of genes is arbitrary. In addition to gaining a deeper understanding of this behavior, exciting future work includes applying cross-attention mechanisms, \textcolor{black}{such as Perceiver-IO~\citep{jaegle2021perceiver} and the MCAB module in scLDM~\citep{palla2025scalable}}. These approaches could learn to optimally aggregate information along the gene dimension of raw expression vectors without relying on the spatial inductive biases that are commonly seen in the convolutional U-Net architectures.

With the emergence of massive single-cell Perturb-seq datasets such as Tahoe-100M~\citep{Zhang2025} and Illumina Billion Cell Atlas~\footnote{\url{https://www.illumina.com/company/news-center/press-releases/2026/fda84c92-b4b3-4691-a402-35555abe8605.html}}, another exciting direction is the development of generative foundation models for virtual cells. Ultimately, this would enable generalization to rare or even unseen contexts and perturbations, either through in-context learning~\citep{dong2026stack} or by leveraging insights from causal inference~\citep{dibaeinia2026virtual}. As both routes can benefit from more perturb-seq data across diverse cellular contexts, having a powerful, high-resolution generative backbone, such as the one established in this paper, will be crucial for driving these models forward.

\textcolor{black}{There are also numerous opportunities to improve the flow matching methodology to better account for biological realism, biological scales, and the practical challenges of modelling single-cell perturb-seq data. 
Ultimately, the transition from modeling virtual cells to simulating entire virtual organisms~\citep{song2024toward}, driven by the principles of multi-modal and multi-scale modeling, may lead to the next major frontier at the intersection of machine learning and biology.}

\bibliography{iclr2026_conference}
\bibliographystyle{iclr2026_conference}

\newpage
\appendix

\newtheorem{dataset}{Dataset}

\section{PerturBench dataset information}
\label{sec:dataset_information_appendix}

In this section, we recapitulate the information of the three main PerturBench benchmarking datasets used in our paper, based on the original information provided in~\cite{wu2024perturbench} section D.1.

\begin{dataset}[\srivatsan{}]
\label{data:Srivatsan20}
This dataset is originally curated from \cite{Srivatsan2020Jan}, featuring 188 small molecule perturbations across the K562, A549, and MCF-7 cell lines, and filtered to contain only the perturbation with the highest dose. In each of the three cell lines, 30\% of perturbations are selected and assigned to val/test split while ensuring they have been observed in the other two cell lines.
\end{dataset}

\begin{dataset}[\jiang{}]
\label{data:Jiang24}
This dataset is originally curated from \cite{Jiang2024Jan}, featuring 219 genetic perturbations across 6 cell lines and 5 cytokine treatments. From the 30 distinct
biological states (6 cell lines $\times$ 5 cytokine treatments), we selected 12 states and held out 70\% of the
perturbations within them for validation and testing.
\end{dataset}

\begin{dataset}[\norman{}]
\label{data:Norman19}
This datasets is originally curated from \citet{norman2019exploring}, featuring 287 gene overexpression perturbations where 131 are dual perturbations in K562 cells. 
\end{dataset}

\section{\textcolor{black}{ARC Virtual Cell Challenge: additional information}}
\label{sec:app_arc_vcc}




\subsection{ARC VCC evaluation metrics}
\label{sec:metrics}

The ARC Virtual Cell Challenge has evaluated model performance using a comprehensive suite of seven metrics. In this section, we detail the three core metrics (DES, PDS, and MAE), as well as the four additional metrics that were introduced in the expanded evaluation suite for the Generalist leaderboard.

\begin{itemize}
    \item \textbf{Differential Expression Score (DES):} DES evaluates how accurately a model predicts differential gene expression. Let $G_{k, true}$ be the ground-truth differentially expressed genes for perturbation $k$, $G_{k, pred}$ be the differentially expressed genes from the model prediction for perturbation $k$, and $\Tilde{G}_{k, pred}$ be the subset of $G_{k, pred}$ where its size is capped to $|\Tilde{G}_{k, pred}|=|G_{k, true}|$ by selecting genes the largest absolute values of log fold changes, then:

    \begin{equation}
        DES_k =
        \begin{cases}
        \frac{|G_{k, pred} \cap G_{k, true}|}{|G_{k, true}|}, & \text{if } |G_{k, pred}| \leq |G_{k, true}| \\
        \frac{|\Tilde{G}_{k, pred} \cap G_{k, true}|}{|G_{k, true}|} & \text{if } |G_{k, pred}| > |G_{k, true}|
        \end{cases}
    \end{equation}

    \textbf{Scale:} 0: lowest possible recall of ground-truth DEGs; 1: highest possible recall of ground-truth DEGs.

    \item \textbf{Perturbation Discrimination Score (PDS):} PDS measures a model's ability to capture the unique effects of a perturbation relative to all other perturbations. This metric is related to the rank metric introduced in PerturBench~\citep{wu2024perturbench}, which, for a given observed perturbation, computes how close the model prediction is to the observation, compared to predictions made for other perturbations. PDS is calculated as:

    \begin{equation}
        PDS_p = 1 - \frac{argind\{d_{pt}\}_{t=p}-1}{N}
    \end{equation}

    where $d_{pt} = d_{L1}(\hat{\delta}_p) - \delta_t |_{\text{sort by t}}$, with $\delta_k = x_k - x_{ntc}$ being the perturbation delta of the pseudobulk expression profile between the perturbed states ($x_k$) and non-targeting controls ($x_{ntc}$).

    \textbf{Scale:} 0: the predicted perturbation effects are not unique to each perturbation; 1: the predicted perturbation effects are highly unique to each perturbation.

    \item \textbf{Mean Absolute Error (MAE):} MAE captures the overall predictive accuracy across the entire gene expression profile. If $G$ is the number of genes, indexed by $g$, the MAE is given by:

    \begin{equation}
        MAE_k = \frac{1}{G} \Sigma_{g=1}^G|\hat{y}_{kg} - y_{kg}|
    \end{equation}

    \textbf{Scale:} 0: perfect fidelity in the predicted gene expression profiles; $\infty$: a complete lack of fidelity.

\end{itemize}

Later into the competition, four other scores were included to for the basis of a Generalist leaderboard:
\begin{itemize}
    \item \textbf{Pearson Delta Correlation:} Measures the correlation between the predicted changes in gene expression and the actual observed changes.

    \textbf{Scale:} -1: a perfectly negative correlation between the predicted and observed changes; 1: a perfectly positive correlation between the predicted and observed changes.

    \item \textbf{Spearman Correlation of Log-Fold Change:} Evaluates how well the ranking of genes from most upregulated to most downregulated matches between the prediction and ground truth, ignoring exact numerical values.

    \textbf{Scale:} -1: completely reversed rankings; 1: perfect match between predicted and observed rankings of genes .

    \item \textbf{Area Under Precision Recall Curve (AUPRC):} A single score summarizing how well a model identifies "true" hits (e.g., differentially expressed genes) while minimizing false positives, which is especially useful when real hits are rare.

    \textbf{Scale:} 0: worst performance; 1: best performance.

    \item \textbf{Spearman Correlation of Effect Size:} Measures the rank consistency of the magnitude of the biological effect (often the t-statistic or Cohen's d) between the predicted and observed data.

    \textbf{Scale:} -1: completely reversed rankings; 1: perfectly matching rankings.
\end{itemize}

\subsection{Data composition and pretraining-finetuning strategies}

\textbf{Pretraining data atlas composition}

Our perturbation data atlas for PRiMeFlow pretraining is composed as follows:
\begin{itemize}
    \item \citet{roohani2025virtual}: The dataset provided by the ARC Virtual Cell Challenge, which includes 150 gene knockdowns as well as non-targeting controls in H1 hESCs, within the training partition.
    \item \citet{replogle2022mapping}: A comprehensive genome-scale Perturb-seq atlas targeting approximately 11,000 genes in K562 and RPE1 cell lines.
    \item \citet{nadig2025transcriptome}: Large-scale Perturb-seq datasets profiling essential gene knockdowns in Jurkat and HepG2 cell lines.
    \item \citet{mcfaline2024multiplex}: A highly multiplexed single-cell chemical genomics (sci-Plex-GxE) dataset probing 522 human kinases combined with targeted drug treatments in glioblastoma cells.
    \item \citet{replogle2020combinatorial}: A small Perturb-seq dataset in K562 cells. We only select the subset of single gene knockdowns for our data atlas.
    \item \citet{Jiang2024Jan}: The same \jiang{} dataset in PerturBench which we have used for benchmarking.
    \item \citet{feng2026genome}: A large Perturb-seq dataset of 7,226 gene knockdowns across 34 iPSC lines from 24 donors.
    \item Internally generated proprietary data, derived from fully differentiated, non-pluripotent cell lines that are biologically distinct from H1 hESCs and iPSCs.
\end{itemize}

In total, our pretraining data atlas comprises 6,975,767 cells spanning seven public datasets and one proprietary dataset. The public data accounts for 89\% of the entire atlas and encompasses 14 distinct cell lines~\footnote{All 14 cell lines from public data: iPSCs, H1 hESCs, HepG2, Jurkat, A549, BxPC-3, HAP1, HT-29, K562, MCF-7, A172, T98G, U-87 MG, and RPE1.} and 10,944 gene knockdowns.





\textbf{Internal validation split on H1 hESCs}
\begin{itemize}
    \item To ensure robust model selection, we established an internal validation split by holding out 43 of the 150 designated training perturbations in H1 hESCs~\footnote{43 held-out perturbations for internal validation: CHMP3, SHPRH, TMSB4X, TCF3, DHCR24, SMARCA5, SSBP1, PBX1, EID2, SRC, HSBP1, MED13, TET1, KDR, TFAM, MAST2, SMAGP, CAST, MTA1, GSK3B, STAT6, ARID1A, CREG1, RRM1, ZNF714, CASP2, CENPO, PTPN1, UBE3C, TGFBR2, MED24, MKI67, TAF13, NRAS, MED1, NIT1, ZNF562, PHF10, WSB2, SIN3B, METTL3, SEC62, BRD9}.
    \item Our internal validation split also includes a subset of 9,544 H1 hESC control cells.
\end{itemize}

\textbf{Finetuning and covariate transfer strategy}

The core objective of our finetuning strategy is to facilitate transfer learning, allowing the model to project perturbation effects learned from the expansive external data atlas onto the specific public and private test perturbations in H1 hESCs.
\begin{itemize}
    \item During the finetuning phase, the validation split is significantly reduced: only 5~\footnote{Validation perturbations at the finetuning stage: EID2, TFAM, PHF10, BRD9, and SIN3B} of the original 43 held-out perturbations are retained for validation.
    \item The remaining 38 perturbations are reassigned back into the training split to maximize the model's exposure to training signals in H1 hESCs.
\end{itemize}

We implemented two distinct finetuning variants to evaluate the impact of targeted data subsets:
\begin{itemize}
    \item PRiMeFlow-300pts finetuning: This dataset includes all 150 training set perturbations sourced from both the VCC H1 hESCs and all external datasets. Furthermore, it incorporates an additional 200 perturbations (comprising 50 public test perturbations and 100 private test perturbations) derived exclusively from the external datasets.
    \item PRiMeFlow-250pts finetuning: This dataset similarly begins with all 150 training set perturbations from both the VCC H1 hESCs and external datasets. However, it isolates the additional data to strictly include the 100 private test perturbations from the external datasets. By narrowing the scope of the included perturbations, this strategy allows the model to focus more on predicting the private test set.
\end{itemize}

Note that we keep all control cells from external data during finetuning.

\subsection{Inference-time strategy}

To generate predictions from PRiMeFlow, we solve the initial value problem using the learned velocity field via the Dopri5 (i.e., the Dormand-Prince method) solver. The ODE integration starts from samples from the standard Gaussian distribution and is conditioned on the specific public or private test perturbation in the H1 hESC cell-line covariate.

During generation, we apply classifier-free guidance with a weight of 20 for perturbed cells and a weight of 5 for synthetic control cells. Empirical evaluations on our internal validation split demonstrated that these asymmetric CFG settings significantly benefit the PDS metric by amplifying the uniqueness and specificity of the predicted perturbation delta. For cell counts, we generate the exact number of perturbed cells recommended by the VCC guidelines (i.e., \textit{pert\_counts\_Validation.csv} and \textit{pert\_counts\_Test.csv}), while maintaining a fixed number of 200 synthetic control cells.

Afterwards, we perform per-perturbation expression normalization, also using the recommended median sequencing depth from the VCC guidelines (\textit{median\_umi\_per\_cell}). Because PRiMeFlow is trained on log1p-normalized data, we execute a postprocessing step:
\begin{enumerate}
    \item convert the predictions to the raw count space through expm1;
    \item perform median normalization;
    \item rounding to the nearest integer;
    \item perform a second median normalization, still using the same recommended median;
    \item convert the expression values back to log-1p space.
\end{enumerate}

Finally, we manually set the target gene expression values in the predicted perturbed expression profiles to 0s.

\section{PRiMeFlow results in ARC VCC public test}
\label{sec:vcc_public_test_ret}

\begin{table}[h]
\caption{\textcolor{black}{PRiMeFlow performance on H1 hESCs public test set using all seven ARC VCC metrics.}}
\centering
\resizebox{\textwidth}{!}{
\begin{tabular}{@{}lccccccc@{}}
\toprule
\multirow{1}{*}{Model} & DES $\uparrow$ & PDS $\uparrow$ & MAE $\downarrow$ & SPEARMAN $\uparrow$ & SPEARMAN\_LFC $\uparrow$ & AUPRC $\uparrow$ & PEARSON $\uparrow$ \\
\toprule
PRiMeFlow pretrained & 0.221 & 0.712 & 0.078 & 0.440 & 0.310 & 0.284 & 0.199 \\
PRiMeFlow 200pts & 0.224 & 0.817 & 0.092 & 0.603 & 0.386 & 0.289 & 0.216 \\
\midrule
x.Compass & 0.344 & 0.852 & 4.215 & - & - & - & - \\
Outlier & 0.378 & 0.832 & 4.218 & - & - & - & - \\
WIND & 0.333 & 0.822 & 0.605 & - & - & - & - \\
\bottomrule
\end{tabular}}
\label{tab:arc_vcc_results_public_test}
\end{table}

Table~\ref{tab:arc_vcc_results_public_test} shows the PRiMeFlow performance in the 50 held-out public test perturabtions in H1 hESCs. The PRiMeFlow-200pts is funetuned to the subset of the pretraining data atlas that include the 250 VCC perturbations (150 training+50 public test) across the H1 hESCs and other external datasets.

We also include the top-three performing models from the ARC VCC public test leaderboard for comparison. Note that we did not include the expanded evaluation metrics for these modes as the public test leaderboard only shows the three core evaluation metrics: DES, PDS and MAE. 

Again, finetuning has been shown to benefit PRiMeFlow across most of the expanded evaluation suite, and overall, PRiMeFlow is able to achieve strong DES and PDS performance, while incurring minimal extortion to the ground truth gene expression values as shown by its significantly lower MAE.

\section{Additional PerturBench results}

\begin{table}[ht]
\caption{Performance of PRiMeFlow on traditional pseudobulk-based metrics in \srivatsan{} dataset.}
\centering
\resizebox{\textwidth}{!}{
\begin{tabular}{@{}lcccc@{}}
\toprule
\multirow{2}{*}{Model} & Cosine & Cosine & RMSE & RMSE \\
                       & logFC & logFC rank & mean & mean rank \\
 \toprule
PRiMeFlow CFG1 & $0.46 \pm 1 \times 10^{-2}$ & $0.12 \pm 1 \times 10^{-2}$ & $0.019 \pm 5 \times 10^{-4}$ & $0.14 \pm 6 \times 10^{-3}$ \\
PRiMeFlow CFG3 & $0.43 \pm 9 \times 10^{-3}$ & $\mathbf{0.10 \pm 4 \times 10^{-3}}$ & $0.029 \pm 3 \times 10^{-4}$ & $0.22 \pm 5 \times 10^{-3}$ \\
PRiMeFlow CFG5 & $0.39 \pm 1 \times 10^{-2}$ & $\mathbf{0.10 \pm 2 \times 10^{-3}}$ & $0.041 \pm 4 \times 10^{-4}$ & $0.29 \pm 1 \times 10^{-2}$ \\
PRiMeFlow MLP CFG1 & $-0.012 \pm 4 \times 10^{-4}$ & $0.46 \pm 1 \times 10^{-2}$ & $0.31 \pm 9 \times 10^{-4}$ & $0.42 \pm 1 \times 10^{-2}$ \\
PRiMeFlow MLP CFG3 & $-0.0071 \pm 2 \times 10^{-3}$ & $0.45 \pm 3 \times 10^{-2}$ & $0.33 \pm 5 \times 10^{-3}$ & $0.44 \pm 2 \times 10^{-2}$ \\
PRiMeFlow MLP CFG5 & $-0.0026 \pm 3 \times 10^{-3}$ & $0.45 \pm 3 \times 10^{-2}$ & $0.40 \pm 1 \times 10^{-2}$ & $0.47 \pm 2 \times 10^{-2}$ \\
\midrule
FM PCA CFG1 & $0.15 \pm 2 \times 10^{-3}$ & $0.34 \pm 6 \times 10^{-3}$ & $0.042 \pm 4 \times 10^{-4}$ & $0.24 \pm 9 \times 10^{-3}$ \\
FM PCA CFG3 & $0.16 \pm 3 \times 10^{-3}$ & $0.30 \pm 3 \times 10^{-3}$ & $0.041 \pm 2 \times 10^{-4}$ & $0.22 \pm 5 \times 10^{-3}$ \\
FM PCA CFG5 & $0.17 \pm 3 \times 10^{-3}$ & $0.28 \pm 6 \times 10^{-3}$ & $0.045 \pm 2 \times 10^{-4}$ & $0.25 \pm 3 \times 10^{-3}$ \\
FM PCA OT CFG1 & $0.12 \pm 2 \times 10^{-4}$ & $0.42 \pm 4 \times 10^{-2}$ & $0.044 \pm 5 \times 10^{-4}$ & $0.41 \pm 2 \times 10^{-2}$ \\
FM PCA OT CFG3 & $0.12 \pm 6 \times 10^{-4}$ & $0.42 \pm 4 \times 10^{-2}$ & $0.044 \pm 5 \times 10^{-4}$ & $0.40 \pm 2 \times 10^{-2}$ \\
FM PCA OT CFG5 & $0.12 \pm 1 \times 10^{-3}$ & $0.42 \pm 4 \times 10^{-2}$ & $0.044 \pm 4 \times 10^{-4}$ & $0.40 \pm 2 \times 10^{-2}$ \\
\midrule
CPA & $0.38 \pm 5 \times 10^{-3}$ & $0.15 \pm 9 \times 10^{-3}$ & $0.020 \pm 2 \times 10^{-4}$ & $0.16 \pm 7 \times 10^{-3}$ \\
CPA (noAdv) & $0.40 \pm 5 \times 10^{-3}$ & $\mathbf{0.09 \pm 5 \times 10^{-3}}$ & $0.020 \pm 1 \times 10^{-4}$ & $\mathbf{0.10 \pm 6 \times 10^{-3}}$ \\
SAMS-VAE & $0.44 \pm 1 \times 10^{-3}$ & $0.17 \pm 1 \times 10^{-2}$ & $0.023 \pm 8 \times 10^{-5}$ & $0.17 \pm 1 \times 10^{-2}$ \\
SAMS-VAE (S) & $\mathbf{0.53 \pm 1 \times 10^{-2}}$ & $0.12 \pm 1 \times 10^{-2}$ & $\mathbf{0.018 \pm 3 \times 10^{-4}}$ & $0.13 \pm 1 \times 10^{-2}$ \\
Biolord & $0.26 \pm 2 \times 10^{-1}$ & $0.26 \pm 2 \times 10^{-1}$ & $0.074 \pm 5 \times 10^{-2}$ & $0.25 \pm 2 \times 10^{-1}$ \\
\midrule
Latent & $0.45 \pm 2 \times 10^{-3}$ & $0.13 \pm 4 \times 10^{-3}$ & $\mathbf{0.018 \pm 6 \times 10^{-5}}$ & $0.15 \pm 3 \times 10^{-3}$ \\
Decoder & $0.35 \pm 5 \times 10^{-3}$ & $0.16 \pm 1 \times 10^{-2}$ & $\mathbf{0.018 \pm 1 \times 10^{-4}}$ & $0.14 \pm 7 \times 10^{-3}$ \\
Linear & $0.16 \pm 1 \times 10^{-2}$ & $0.28 \pm 5 \times 10^{-3}$ & $0.030 \pm 5 \times 10^{-4}$ & $0.27 \pm 2 \times 10^{-3}$ \\
\bottomrule
\end{tabular}}
\label{tab:srivatsan_additional_pseudo}
\end{table}

\begin{table}[ht]
\caption{Performance of PRiMeFlow on traditional pseudobulk-based metrics in \norman{} dataset.}
\centering
\resizebox{\textwidth}{!}{
\begin{tabular}{@{}lcccc@{}}
\toprule
\multirow{2}{*}{Model} & Cosine & Cosine & RMSE & RMSE \\
                       & logFC & logFC rank & mean & mean rank \\
 \toprule
PRiMeFlow CFG1 & $0.77 \pm 4 \times 10^{-3}$ & $0.0057 \pm 8 \times 10^{-4}$ & $0.046 \pm 1 \times 10^{-3}$ & $0.022 \pm 6 \times 10^{-3}$ \\
PRiMeFlow CFG3 & $\mathbf{0.79 \pm 2 \times 10^{-3}}$ & $0.0072 \pm 3 \times 10^{-3}$ & $0.074 \pm 2 \times 10^{-3}$ & $0.077 \pm 8 \times 10^{-3}$ \\
PRiMeFlow CFG5 & $0.74 \pm 3 \times 10^{-3}$ & $0.015 \pm 3 \times 10^{-3}$ & $0.11 \pm 2 \times 10^{-3}$ & $0.19 \pm 1 \times 10^{-2}$ \\
PRiMeFlow MLP CFG1 & $0.024 \pm 6 \times 10^{-3}$ & $0.27 \pm 3 \times 10^{-2}$ & $0.29 \pm 2 \times 10^{-3}$ & $0.16 \pm 4 \times 10^{-2}$ \\
PRiMeFlow MLP CFG3 & $0.054 \pm 8 \times 10^{-3}$ & $0.27 \pm 2 \times 10^{-2}$ & $0.31 \pm 2 \times 10^{-3}$ & $0.25 \pm 2 \times 10^{-2}$ \\
PRiMeFlow MLP CFG5 & $0.080 \pm 9 \times 10^{-3}$ & $0.26 \pm 2 \times 10^{-2}$ & $0.33 \pm 4 \times 10^{-3}$ & $0.31 \pm 1 \times 10^{-2}$ \\
\midrule
FM PCA CFG1 & $0.50 \pm 2 \times 10^{-3}$ & $0.053 \pm 3 \times 10^{-3}$ & $0.062 \pm 9 \times 10^{-5}$ & $0.067 \pm 3 \times 10^{-3}$ \\
FM PCA CFG3 & $0.61 \pm 3 \times 10^{-3}$ & $0.040 \pm 2 \times 10^{-3}$ & $0.076 \pm 6 \times 10^{-4}$ & $0.083 \pm 1 \times 10^{-3}$ \\
FM PCA CFG5 & $0.63 \pm 4 \times 10^{-3}$ & $0.034 \pm 3 \times 10^{-3}$ & $0.11 \pm 2 \times 10^{-3}$ & $0.15 \pm 1 \times 10^{-2}$ \\
FM PCA OT CFG1 & $0.056 \pm 5 \times 10^{-3}$ & $0.32 \pm 2 \times 10^{-2}$ & $0.095 \pm 7 \times 10^{-4}$ & $0.33 \pm 2 \times 10^{-2}$ \\
FM PCA OT CFG3 & $0.060 \pm 6 \times 10^{-3}$ & $0.32 \pm 2 \times 10^{-2}$ & $0.094 \pm 8 \times 10^{-4}$ & $0.32 \pm 2 \times 10^{-2}$ \\
FM PCA OT CFG5 & $0.065 \pm 7 \times 10^{-3}$ & $0.32 \pm 2 \times 10^{-2}$ & $0.094 \pm 8 \times 10^{-4}$ & $0.32 \pm 2 \times 10^{-2}$ \\
\midrule
CPA & $0.76 \pm 4 \times 10^{-3}$ & $0.0067 \pm 2 \times 10^{-3}$ & $0.046 \pm 4 \times 10^{-4}$ & $0.019 \pm 3 \times 10^{-3}$ \\
CPA (noAdv) & $0.77 \pm 1 \times 10^{-2}$ & $\mathbf{0.0053 \pm 3 \times 10^{-3}}$ & $0.046 \pm 1 \times 10^{-3}$ & $0.017 \pm 1 \times 10^{-3}$ \\
SAMS-VAE & $0.45 \pm 2 \times 10^{-2}$ & $0.019 \pm 7 \times 10^{-3}$ & $0.084 \pm 7 \times 10^{-4}$ & $0.025 \pm 4 \times 10^{-3}$ \\
SAMS-VAE (S) & $\mathbf{0.78 \pm 6 \times 10^{-3}}$ & $0.020 \pm 4 \times 10^{-3}$ & $0.047 \pm 2 \times 10^{-3}$ & $0.030 \pm 6 \times 10^{-3}$ \\
Biolord & $0.45 \pm 8 \times 10^{-3}$ & $0.026 \pm 4 \times 10^{-3}$ & $0.083 \pm 2 \times 10^{-4}$ & $0.034 \pm 2 \times 10^{-3}$ \\
\midrule
Latent & $\mathbf{0.79 \pm 1 \times 10^{-2}}$ & $\mathbf{0.005 \pm 2 \times 10^{-3}}$ & $\mathbf{0.043 \pm 4 \times 10^{-4}}$ & $\mathbf{0.014 \pm 1 \times 10^{-3}}$ \\
Decoder & $0.73 \pm 2 \times 10^{-2}$ & $0.017 \pm 6 \times 10^{-3}$ & $\mathbf{0.043 \pm 3 \times 10^{-4}}$ & $\mathbf{0.014 \pm 4 \times 10^{-4}}$ \\
Linear & $0.60 \pm 2 \times 10^{-2}$ & $0.035 \pm 4 \times 10^{-3}$ & $0.057 \pm 3 \times 10^{-3}$ & $0.016 \pm 8 \times 10^{-4}$ \\
\bottomrule
\end{tabular}}
\label{tab:norman_additional_pseudo}
\end{table}

\begin{table}[ht]
\caption{Performance of PRiMeFlow on traditional pseudobulk-based metrics in \jiang{} dataset.}
\centering
\resizebox{\textwidth}{!}{
\begin{tabular}{lllll}
\toprule
 \multirow{2}{*}{Model} & Cosine & Cosine & RMSE & RMSE \\
                       & logFC & logFC rank & mean & mean rank \\
\toprule
PRiMeFlow CFG1 & $0.47 \pm 5 \times 10^{-3}$ & $0.39 \pm 2 \times 10^{-2}$ & $0.019 \pm 3 \times 10^{-4}$ & $0.40 \pm 2 \times 10^{-2}$ \\
PRiMeFlow CFG3 & $0.41 \pm 4 \times 10^{-3}$ & $0.38 \pm 1 \times 10^{-2}$ & $0.034 \pm 5 \times 10^{-4}$ & $0.43 \pm 1 \times 10^{-2}$ \\
PRiMeFlow CFG5 & $0.36 \pm 4 \times 10^{-3}$ & $0.37 \pm 1 \times 10^{-2}$ & $0.045 \pm 5 \times 10^{-4}$ & $0.43 \pm 7 \times 10^{-3}$ \\
PRiMeFlow MLP CFG1 & $0.053 \pm 1 \times 10^{-4}$ & $0.45 \pm 5 \times 10^{-3}$ & $0.33 \pm 6 \times 10^{-4}$ & $0.46 \pm 1 \times 10^{-3}$ \\
PRiMeFlow MLP CFG3 & $0.053 \pm 7 \times 10^{-5}$ & $0.43 \pm 1 \times 10^{-2}$ & $0.33 \pm 6 \times 10^{-4}$ & $0.46 \pm 3 \times 10^{-3}$ \\
PRiMeFlow MLP CFG5 & $0.053 \pm 9 \times 10^{-5}$ & $0.43 \pm 1 \times 10^{-2}$ & $0.33 \pm 7 \times 10^{-4}$ & $0.47 \pm 3 \times 10^{-3}$ \\
\midrule
FlowMatching PCA CFG1 & $0.071 \pm 8 \times 10^{-5}$ & $0.37 \pm 9 \times 10^{-3}$ & $0.10 \pm 0$ & $0.39 \pm 9 \times 10^{-3}$ \\
FlowMatching PCA CFG3 & $0.072 \pm 1 \times 10^{-4}$ & $0.36 \pm 4 \times 10^{-3}$ & $0.10 \pm 9 \times 10^{-5}$ & $0.42 \pm 5 \times 10^{-3}$ \\
FlowMatching PCA CFG5 & $0.072 \pm 3 \times 10^{-4}$ & $0.36 \pm 5 \times 10^{-3}$ & $0.10 \pm 2 \times 10^{-4}$ & $0.44 \pm 5 \times 10^{-3}$ \\
FlowMatching PCA OT CFG1 & $0.068 \pm 3 \times 10^{-4}$ & $0.47 \pm 2 \times 10^{-2}$ & $0.10 \pm 8 \times 10^{-4}$ & $0.47 \pm 4 \times 10^{-3}$ \\
FlowMatching PCA OT CFG3 & $0.068 \pm 4 \times 10^{-4}$ & $0.47 \pm 2 \times 10^{-2}$ & $0.11 \pm 2 \times 10^{-4}$ & $0.47 \pm 8 \times 10^{-4}$ \\
FlowMatching PCA OT CFG5 & $0.068 \pm 6 \times 10^{-4}$ & $0.47 \pm 1 \times 10^{-2}$ & $0.11 \pm 2 \times 10^{-3}$ & $0.47 \pm 3 \times 10^{-3}$ \\
\midrule
CPA & $0.60 \pm 2 \times 10^{-3}$ & $0.40 \pm 9 \times 10^{-3}$ & $\mathbf{0.015 \pm 5 \times 10^{-5}}$ & $0.41 \pm 1 \times 10^{-2}$ \\
CPA (noAdv) & $0.60 \pm 2 \times 10^{-3}$ & $0.39 \pm 1 \times 10^{-2}$ & $\mathbf{0.015 \pm 4 \times 10^{-5}}$ & $0.40 \pm 1 \times 10^{-2}$ \\
SAMS-VAE & $0.59 \pm 3 \times 10^{-3}$ & $0.43 \pm 1 \times 10^{-2}$ & $0.017 \pm 1 \times 10^{-4}$ & $0.43 \pm 1 \times 10^{-2}$ \\
SAMS-VAE (S) & $0.57 \pm 5 \times 10^{-2}$ & $0.43 \pm 1 \times 10^{-2}$ & $0.017 \pm 2 \times 10^{-3}$ & $0.42 \pm 4 \times 10^{-2}$ \\
\midrule
Latent & $0.47 \pm 1 \times 10^{-3}$ & $0.38 \pm 6 \times 10^{-3}$ & $\mathbf{0.015 \pm 6 \times 10^{-5}}$ & $0.38 \pm 7 \times 10^{-3}$ \\
Decoder & $\mathbf{0.64 \pm 1 \times 10^{-3}}$ & $\mathbf{0.32 \pm 8 \times 10^{-3}}$ & $\mathbf{0.015 \pm 3 \times 10^{-5}}$ & $\mathbf{0.32 \pm 5 \times 10^{-3}}$ \\
Linear & $0.17 \pm 9 \times 10^{-5}$ & $0.34 \pm 2 \times 10^{-4}$ & $0.038 \pm 5 \times 10^{-5}$ & $0.43 \pm 1 \times 10^{-3}$ \\
\bottomrule
\end{tabular}}
\label{tab:jiang_additional_pseudo}
\end{table}

\clearpage

\begin{sidewaystable}
\caption{All distributional metrics performance on the \srivatsan{} dataset.}
\begin{tabular}{@{}lccccccc@{}}
\toprule
\multirow{2}{*}{Model} & MMD & MMD & MMD & MMD & DEG & DEG \\
                       & GEX & GEX rank & PCA & PCA rank & recall & recall rank\\
\toprule
PRiMeFlow CFG1 & $\mathbf{0.13 \pm 4 \times 10^{-3}}$ & $\mathbf{0.13 \pm 8 \times 10^{-3}}$ & $\mathbf{0.14 \pm 8 \times 10^{-3}}$ & $0.14 \pm 9 \times 10^{-3}$ & $\mathbf{0.26 \pm 1 \times 10^{-2}}$ & $0.27 \pm 2 \times 10^{-2}$ \\
PRiMeFlow CFG3 & $0.25 \pm 2 \times 10^{-3}$ & $0.22 \pm 7 \times 10^{-3}$ & $0.31 \pm 1 \times 10^{-2}$ & $0.21 \pm 9 \times 10^{-3}$ & $\mathbf{0.27 \pm 6 \times 10^{-3}}$ & $\mathbf{0.24 \pm 1 \times 10^{-2}}$ \\
PRiMeFlow CFG5 & $0.47 \pm 1 \times 10^{-2}$ & $0.29 \pm 2 \times 10^{-2}$ & $0.60 \pm 2 \times 10^{-2}$ & $0.26 \pm 9 \times 10^{-3}$ & $0.25 \pm 5 \times 10^{-3}$ & $\mathbf{0.24 \pm 1 \times 10^{-2}}$ \\
PRiMeFlow MLP CFG1 & $14. \pm 5 \times 10^{-2}$ & $0.43 \pm 2 \times 10^{-2}$ & $8.8 \pm 1 \times 10^{-1}$ & $0.42 \pm 2 \times 10^{-2}$ & $0.011 \pm 8 \times 10^{-4}$ & $0.47 \pm 6 \times 10^{-3}$ \\
PRiMeFlow MLP CFG3 & $15. \pm 2 \times 10^{-1}$ & $0.44 \pm 2 \times 10^{-2}$ & $11. \pm 6 \times 10^{-1}$ & $0.43 \pm 5 \times 10^{-3}$ & $0.11 \pm 2 \times 10^{-2}$ & $0.49 \pm 2 \times 10^{-2}$ \\
PRiMeFlow MLP CFG5 & $18. \pm 8 \times 10^{-1}$ & $0.47 \pm 2 \times 10^{-2}$ & $18. \pm 1 \times 10^{0}$ & $0.46 \pm 2 \times 10^{-2}$ & $0.13 \pm 1 \times 10^{-2}$ & $0.49 \pm 4 \times 10^{-2}$ \\
\midrule
FM PCA CFG1 & $1.2 \pm 7 \times 10^{-3}$ & $0.26 \pm 6 \times 10^{-3}$ & $0.50 \pm 2 \times 10^{-2}$ & $0.24 \pm 1 \times 10^{-2}$ & $0.042 \pm 2 \times 10^{-3}$ & $0.46 \pm 6 \times 10^{-3}$ \\
FM PCA CFG3 & $1.3 \pm 5 \times 10^{-3}$ & $0.25 \pm 7 \times 10^{-3}$ & $0.44 \pm 1 \times 10^{-2}$ & $0.22 \pm 8 \times 10^{-3}$ & $0.071 \pm 3 \times 10^{-3}$ & $0.39 \pm 1 \times 10^{-2}$ \\
FM PCA CFG5 & $1.4 \pm 4 \times 10^{-3}$ & $0.29 \pm 4 \times 10^{-3}$ & $0.53 \pm 1 \times 10^{-2}$ & $0.26 \pm 6 \times 10^{-3}$ & $0.10 \pm 3 \times 10^{-3}$ & $0.36 \pm 9 \times 10^{-3}$ \\
FM PCA OT CFG1 & $1.3 \pm 3 \times 10^{-2}$ & $0.43 \pm 2 \times 10^{-2}$ & $0.62 \pm 5 \times 10^{-3}$ & $0.40 \pm 3 \times 10^{-2}$ & $0.023 \pm 4 \times 10^{-4}$ & $0.51 \pm 3 \times 10^{-3}$ \\
FM PCA OT CFG3 & $1.3 \pm 3 \times 10^{-2}$ & $0.42 \pm 2 \times 10^{-2}$ & $0.61 \pm 4 \times 10^{-3}$ & $0.40 \pm 3 \times 10^{-2}$ & $0.024 \pm 6 \times 10^{-4}$ & $0.51 \pm 2 \times 10^{-3}$ \\
FM PCA OT CFG5 & $1.3 \pm 3 \times 10^{-2}$ & $0.43 \pm 2 \times 10^{-2}$ & $0.60 \pm 4 \times 10^{-3}$ & $0.39 \pm 2 \times 10^{-2}$ & $0.024 \pm 6 \times 10^{-4}$ & $0.51 \pm 4 \times 10^{-3}$ \\
\midrule
CPA & $2.4 \pm 1 \times 10^{-2}$ & $0.30 \pm 9 \times 10^{-3}$ & $0.53 \pm 4 \times 10^{-3}$ & $0.20 \pm 9 \times 10^{-3}$ & $0.0073 \pm 2 \times 10^{-3}$ & $0.46 \pm 5 \times 10^{-3}$ \\
CPA (noAdv) & $2.3 \pm 3 \times 10^{-2}$ & $0.25 \pm 5 \times 10^{-3}$ & $0.49 \pm 1 \times 10^{-2}$ & $0.13 \pm 1 \times 10^{-2}$ & $0.0040 \pm 2 \times 10^{-3}$ & $0.48 \pm 5 \times 10^{-2}$ \\
SAMS-VAE & $2.5 \pm 2 \times 10^{-2}$ & $0.30 \pm 8 \times 10^{-3}$ & $0.69 \pm 1 \times 10^{-2}$ & $\mathbf{0.060 \pm 3 \times 10^{-3}}$ & $0 \pm 0$ & $0.50 \pm 4 \times 10^{-2}$ \\ 
SAMS-VAE (S) & $2.9 \pm 1 \times 10^{-2}$ & $0.28 \pm 5 \times 10^{-3}$ & $0.79 \pm 1 \times 10^{-2}$ & $0.18 \pm 2 \times 10^{-2}$ & $0 \pm 0$ & $0.50 \pm 4 \times 10^{-2}$ \\
Biolord & $4.9 \pm 3 \times 10^{0}$ & $0.36 \pm 2 \times 10^{-1}$ & $4.3 \pm 4 \times 10^{0}$ & $0.32 \pm 2 \times 10^{-1}$ & $4.7e-05 \pm 1 \times 10^{-4}$ & $0.49 \pm 2 \times 10^{-2}$ \\
\midrule
Latent & $4.3 \pm 2 \times 10^{-1}$ & $0.26 \pm 6 \times 10^{-2}$ & $2.0 \pm 2 \times 10^{-1}$ & $0.26 \pm 5 \times 10^{-2}$ & $0.0 \pm 0.0 $ & $0.49 \pm 3 \times 10^{-2}$ \\
Decoder & $4.2 \pm 4 \times 10^{-3}$ & $0.16 \pm 2 \times 10^{-2}$ & $1.9 \pm 5 \times 10^{-3}$ & $0.15 \pm 1 \times 10^{-2}$ & N/A & N/A \\
Linear & $2.2 \pm 7 \times 10^{-3}$ & $0.31 \pm 1 \times 10^{-3}$ & $0.76 \pm 9 \times 10^{-4}$ & $0.30 \pm 3 \times 10^{-4}$ & $0.0036 \pm 3 \times 10^{-4}$ & $0.49 \pm 5 \times 10^{-2}$ \\
\bottomrule
\end{tabular}
\label{tab:srivatsan_full_dist}
\end{sidewaystable}

\clearpage

\begin{sidewaystable}
\caption{All distributional metrics performance on the \norman{} dataset.}
\begin{tabular}{@{}lccccccc@{}}
\toprule
\multirow{2}{*}{Model} & MMD & MMD & MMD & MMD & DEG & DEG \\
                       & GEX & GEX rank & PCA & PCA rank & recall & recall rank\\
\toprule
PRiMeFlow CFG1 & $\mathbf{0.26 \pm 7 \times 10^{-3}}$ & $\mathbf{0.019 \pm 4 \times 10^{-3}}$ & $\mathbf{0.34 \pm 2 \times 10^{-2}}$ & $0.023 \pm 5 \times 10^{-3}$ & $0.54 \pm 2 \times 10^{-2}$ & $0.024 \pm 6 \times 10^{-3}$ \\
PRiMeFlow CFG3 & $0.61 \pm 3 \times 10^{-2}$ & $0.066 \pm 9 \times 10^{-3}$ & $0.86 \pm 5 \times 10^{-2}$ & $0.063 \pm 6 \times 10^{-3}$ & $\mathbf{0.61 \pm 8 \times 10^{-3}}$ & $\mathbf{0.0060 \pm 3 \times 10^{-3}}$ \\
PRiMeFlow CFG5 & $1.3 \pm 4 \times 10^{-2}$ & $0.17 \pm 1 \times 10^{-2}$ & $1.8 \pm 5 \times 10^{-2}$ & $0.14 \pm 1 \times 10^{-2}$ & $0.56 \pm 1 \times 10^{-2}$ & $0.0085 \pm 2 \times 10^{-3}$ \\
PRiMeFlow MLP CFG1 & $7.2 \pm 8 \times 10^{-2}$ & $0.19 \pm 4 \times 10^{-2}$ & $3.7 \pm 1 \times 10^{-1}$ & $0.16 \pm 4 \times 10^{-2}$ & $0.057 \pm 8 \times 10^{-3}$ & $0.42 \pm 2 \times 10^{-2}$ \\
PRiMeFlow MLP CFG3 & $7.6 \pm 9 \times 10^{-2}$ & $0.27 \pm 2 \times 10^{-2}$ & $4.5 \pm 2 \times 10^{-1}$ & $0.25 \pm 2 \times 10^{-2}$ & $0.20 \pm 3 \times 10^{-2}$ & $0.28 \pm 1 \times 10^{-2}$ \\
PRiMeFlow MLP CFG5 & $8.5 \pm 2 \times 10^{-1}$ & $0.32 \pm 8 \times 10^{-3}$ & $6.3 \pm 4 \times 10^{-1}$ & $0.31 \pm 1 \times 10^{-2}$ & $0.30 \pm 3 \times 10^{-2}$ & $0.21 \pm 3 \times 10^{-2}$ \\
\midrule
FM PCA CFG1 & $0.77 \pm 4 \times 10^{-3}$ & $0.059 \pm 3 \times 10^{-3}$ & $0.61 \pm 5 \times 10^{-3}$ & $0.066 \pm 4 \times 10^{-3}$ & $0.37 \pm 2 \times 10^{-3}$ & $0.069 \pm 9 \times 10^{-3}$ \\
FM PCA CFG3 & $1.1 \pm 8 \times 10^{-3}$ & $0.098 \pm 2 \times 10^{-3}$ & $0.82 \pm 9 \times 10^{-3}$ & $0.076 \pm 1 \times 10^{-3}$ & $0.45 \pm 2 \times 10^{-3}$ & $0.030 \pm 2 \times 10^{-3}$ \\
FM PCA CFG5 & $1.9 \pm 4 \times 10^{-2}$ & $0.18 \pm 1 \times 10^{-2}$ & $1.6 \pm 4 \times 10^{-2}$ & $0.12 \pm 4 \times 10^{-3}$ & $0.43 \pm 4 \times 10^{-3}$ & $0.031 \pm 3 \times 10^{-3}$ \\
FM PCA OT CFG1 & $2.6 \pm 1 \times 10^{-2}$ & $0.33 \pm 2 \times 10^{-2}$ & $1.8 \pm 2 \times 10^{-2}$ & $0.32 \pm 2 \times 10^{-2}$ & $0.0012 \pm 9 \times 10^{-4}$ & $0.56 \pm 2 \times 10^{-2}$ \\
FM PCA OT CFG3 & $2.6 \pm 1 \times 10^{-2}$ & $0.32 \pm 2 \times 10^{-2}$ & $1.8 \pm 2 \times 10^{-2}$ & $0.32 \pm 2 \times 10^{-2}$ & $0.0017 \pm 1 \times 10^{-3}$ & $0.55 \pm 2 \times 10^{-2}$ \\
FM PCA OT CFG5 & $2.6 \pm 2 \times 10^{-2}$ & $0.32 \pm 2 \times 10^{-2}$ & $1.8 \pm 3 \times 10^{-2}$ & $0.32 \pm 2 \times 10^{-2}$ & $0.0020 \pm 1 \times 10^{-3}$ & $0.54 \pm 2 \times 10^{-2}$ \\
\midrule
CPA & $5.6 \pm 3 \times 10^{-2}$ & $0.14 \pm 9 \times 10^{-3}$ & $2.2 \pm 2 \times 10^{-2}$ & $0.044 \pm 5 \times 10^{-3}$ & $0.032 \pm 4 \times 10^{-3}$ & $0.30 \pm 5 \times 10^{-2}$ \\
CPA (noAdv) & $5.5 \pm 1 \times 10^{-1}$ & $0.13 \pm 2 \times 10^{-2}$ & $2.2 \pm 1 \times 10^{-1}$ & $0.038 \pm 5 \times 10^{-3}$ & $0.016 \pm 9 \times 10^{-3}$ & $0.28 \pm 5 \times 10^{-2}$ \\
SAMS-VAE & $4.1 \pm 4 \times 10^{-2}$ & $0.034 \pm 1 \times 10^{-2}$ & $1.9 \pm 3 \times 10^{-2}$ & $0.020 \pm 2 \times 10^{-3}$ & $0 \pm 0$ & $0.46 \pm 4 \times 10^{-2}$ \\
SAMS-VAE (S) & $3.3 \pm 5 \times 10^{-2}$ & $0.088 \pm 6 \times 10^{-3}$ & $0.74 \pm 5 \times 10^{-2}$ & $0.029 \pm 4 \times 10^{-3}$ & $0.028 \pm 6 \times 10^{-3}$ & $0.38 \pm 3 \times 10^{-2}$ \\
Biolord & $2.8 \pm 5 \times 10^{-3}$ & $0.066 \pm 3 \times 10^{-3}$ & $1.6 \pm 5 \times 10^{-3}$ & $0.042 \pm 3 \times 10^{-3}$ & $0 \pm 0$ & $0.47 \pm 3 \times 10^{-2}$ \\
\midrule
Latent & $6.7 \pm 4 \times 10^{-3}$ & $0.012 \pm 1 \times 10^{-3}$ & $3.2 \pm 6 \times 10^{-3}$ & $0.013 \pm 8 \times 10^{-4}$ & $0.00043 \pm 3 \times 10^{-4}$ & $0.50 \pm 1 \times 10^{-2}$ \\
Decoder & $6.7 \pm 6 \times 10^{-3}$ & $0.0090 \pm 8 \times 10^{-4}$ & $3.2 \pm 4 \times 10^{-3}$ & $\mathbf{0.0098 \pm 4 \times 10^{-4}}$ & N/A & N/A \\
Linear & $2.5 \pm 4 \times 10^{-2}$ & $0.026 \pm 2 \times 10^{-3}$ & $1.2 \pm 4 \times 10^{-2}$ & $0.017 \pm 4 \times 10^{-4}$ & $0.018 \pm 2 \times 10^{-3}$ & $0.40 \pm 4 \times 10^{-2}$ \\
\bottomrule
\end{tabular}
\label{tab:norman_full_dist}
\end{sidewaystable}

\clearpage

\begin{sidewaystable}
\caption{All distributional metrics performance on the \jiang{} dataset.}
\begin{tabular}{lllllll}
\toprule
 \multirow{2}{*}{Model} & MMD & MMD & MMD & MMD & DEG & DEG \\
                       & GEX & GEX rank & PCA & PCA rank & recall & recall rank\\
\toprule
PRiMeFlow CFG1 & $\mathbf{0.14 \pm 4 \times 10^{-3}}$ & $0.41 \pm 9 \times 10^{-3}$ & $\mathbf{0.057 \pm 6 \times 10^{-4}}$ & $\mathbf{0.39 \pm 1 \times 10^{-2}}$ & $\mathbf{0.099 \pm 5 \times 10^{-3}}$ & $\mathbf{0.41 \pm 3 \times 10^{-2}}$ \\
PRiMeFlow CFG3 & $0.37 \pm 4 \times 10^{-2}$ & $0.44 \pm 1 \times 10^{-2}$ & $0.38 \pm 2 \times 10^{-2}$ & $0.44 \pm 8 \times 10^{-3}$ & $0.061 \pm 4 \times 10^{-3}$ & $0.43 \pm 7 \times 10^{-3}$ \\
PRiMeFlow CFG5 & $0.73 \pm 7 \times 10^{-2}$ & $0.44 \pm 9 \times 10^{-3}$ & $0.73 \pm 3 \times 10^{-2}$ & $0.44 \pm 9 \times 10^{-3}$ & $0.056 \pm 3 \times 10^{-3}$ & $0.43 \pm 2 \times 10^{-2}$ \\
PRiMeFlow MLP CFG1 & $17. \pm 7 \times 10^{-2}$ & $0.46 \pm 2 \times 10^{-3}$ & $9.5 \pm 8 \times 10^{-2}$ & $0.46 \pm 3 \times 10^{-3}$ & $0.0037 \pm 3 \times 10^{-4}$ & $0.45 \pm 1 \times 10^{-2}$ \\
PRiMeFlow MLP CFG3 & $17. \pm 4 \times 10^{-2}$ & $0.46 \pm 3 \times 10^{-3}$ & $9.8 \pm 1 \times 10^{-1}$ & $0.46 \pm 2 \times 10^{-3}$ & $0.0031 \pm 3 \times 10^{-4}$ & $0.44 \pm 4 \times 10^{-3}$ \\
PRiMeFlow MLP CFG5 & $17. \pm 5 \times 10^{-2}$ & $0.47 \pm 3 \times 10^{-3}$ & $9.8 \pm 1 \times 10^{-1}$ & $0.46 \pm 2 \times 10^{-3}$ & $0.0033 \pm 5 \times 10^{-4}$ & $0.45 \pm 1 \times 10^{-2}$ \\
\midrule
FlowMatching PCA CFG1 & $6.2 \pm 9 \times 10^{-3}$ & $0.46 \pm 2 \times 10^{-3}$ & $4.2 \pm 4 \times 10^{-3}$ & $0.45 \pm 3 \times 10^{-3}$ & $0.0055 \pm 0$ & $0.46 \pm 0$ \\
FlowMatching PCA CFG3 & $6.8 \pm 3 \times 10^{-2}$ & $0.46 \pm 3 \times 10^{-3}$ & $4.0 \pm 1 \times 10^{-2}$ & $0.44 \pm 6 \times 10^{-3}$ & $0.0055 \pm 0$ & $0.46 \pm 0$ \\
FlowMatching PCA CFG5 & $7.1 \pm 7 \times 10^{-2}$ & $0.46 \pm 2 \times 10^{-3}$ & $4.1 \pm 3 \times 10^{-2}$ & $0.44 \pm 7 \times 10^{-3}$ & $0.0056 \pm 6 \times 10^{-5}$ & $0.46 \pm 3 \times 10^{-3}$ \\
FlowMatching PCA OT CFG1 & $6.3 \pm 9 \times 10^{-2}$ & $0.47 \pm 9 \times 10^{-4}$ & $3.7 \pm 2 \times 10^{-2}$ & $0.47 \pm 4 \times 10^{-3}$ & $0.0055 \pm 0$ & $0.46 \pm 0$ \\
FlowMatching PCA OT CFG3 & $6.5 \pm 1 \times 10^{-1}$ & $0.47 \pm 3 \times 10^{-3}$ & $3.6 \pm 2 \times 10^{-2}$ & $0.47 \pm 4 \times 10^{-3}$ & $0.0055 \pm 0$ & $0.46 \pm 3 \times 10^{-4}$ \\
FlowMatching PCA OT CFG5 & $6.7 \pm 2 \times 10^{-1}$ & $0.47 \pm 1 \times 10^{-3}$ & $3.6 \pm 4 \times 10^{-2}$ & $0.47 \pm 5 \times 10^{-3}$ & $0.0057 \pm 2 \times 10^{-4}$ & $0.46 \pm 4 \times 10^{-3}$ \\
\midrule
CPA & $8.0 \pm 2 \times 10^{-3}$ & $0.44 \pm 7 \times 10^{-3}$ & $2.5 \pm 4 \times 10^{-3}$ & $0.43 \pm 4 \times 10^{-3}$ & $0.0044 \pm 2 \times 10^{-3}$ & $0.48 \pm 5 \times 10^{-3}$ \\
CPA (noAdv) & $8.0 \pm 2 \times 10^{-3}$ & $0.44 \pm 6 \times 10^{-3}$ & $2.5 \pm 3 \times 10^{-3}$ & $0.42 \pm 4 \times 10^{-3}$ & $0.0052 \pm 6 \times 10^{-4}$ & $0.48 \pm 1 \times 10^{-2}$ \\
SAMS-VAE & $4.5 \pm 7 \times 10^{-3}$ & $0.47 \pm 3 \times 10^{-3}$ & $0.32 \pm 5 \times 10^{-3}$ & $0.43 \pm 1 \times 10^{-2}$ & $1.8e-05 \pm 4 \times 10^{-5}$ & $0.48 \pm 1 \times 10^{-3}$ \\
SAMS-VAE (S) & $6.5 \pm 2 \times 10^{0}$ & $0.44 \pm 2 \times 10^{-2}$ & $1.6 \pm 1 \times 10^{0}$ & $0.43 \pm 1 \times 10^{-2}$ & $0.0015 \pm 1 \times 10^{-4}$ & $0.48 \pm 4 \times 10^{-3}$ \\
\midrule
Latent & $8.0 \pm 9 \times 10^{-4}$ & $0.41 \pm 5 \times 10^{-3}$ & $2.6 \pm 2 \times 10^{-3}$ & $0.40 \pm 5 \times 10^{-3}$ & $0.0010 \pm 5 \times 10^{-4}$ & $0.47 \pm 6 \times 10^{-3}$ \\
Decoder & $8.0 \pm 2 \times 10^{-3}$ & $\mathbf{0.36 \pm 9 \times 10^{-3}}$ & $2.6 \pm 2 \times 10^{-3}$ & $\mathbf{0.38 \pm 8 \times 10^{-3}}$ & N/A & N/A \\
Linear & $3.1 \pm 2 \times 10^{-3}$ & $0.44 \pm 1 \times 10^{-3}$ & $1.3 \pm 2 \times 10^{-3}$ & $0.45 \pm 6 \times 10^{-4}$ & $0.0033 \pm 2 \times 10^{-4}$ & $0.48 \pm 3 \times 10^{-3}$ \\
\bottomrule
\end{tabular}
\label{tab:jiang_full_dist}
\end{sidewaystable}

\clearpage

\section{Additional PCA visualization for \srivatsan{} dataset}
\label{sec:add_pca_sri}

\begin{figure}[h]
\centering
\includegraphics[width=1.\textwidth]{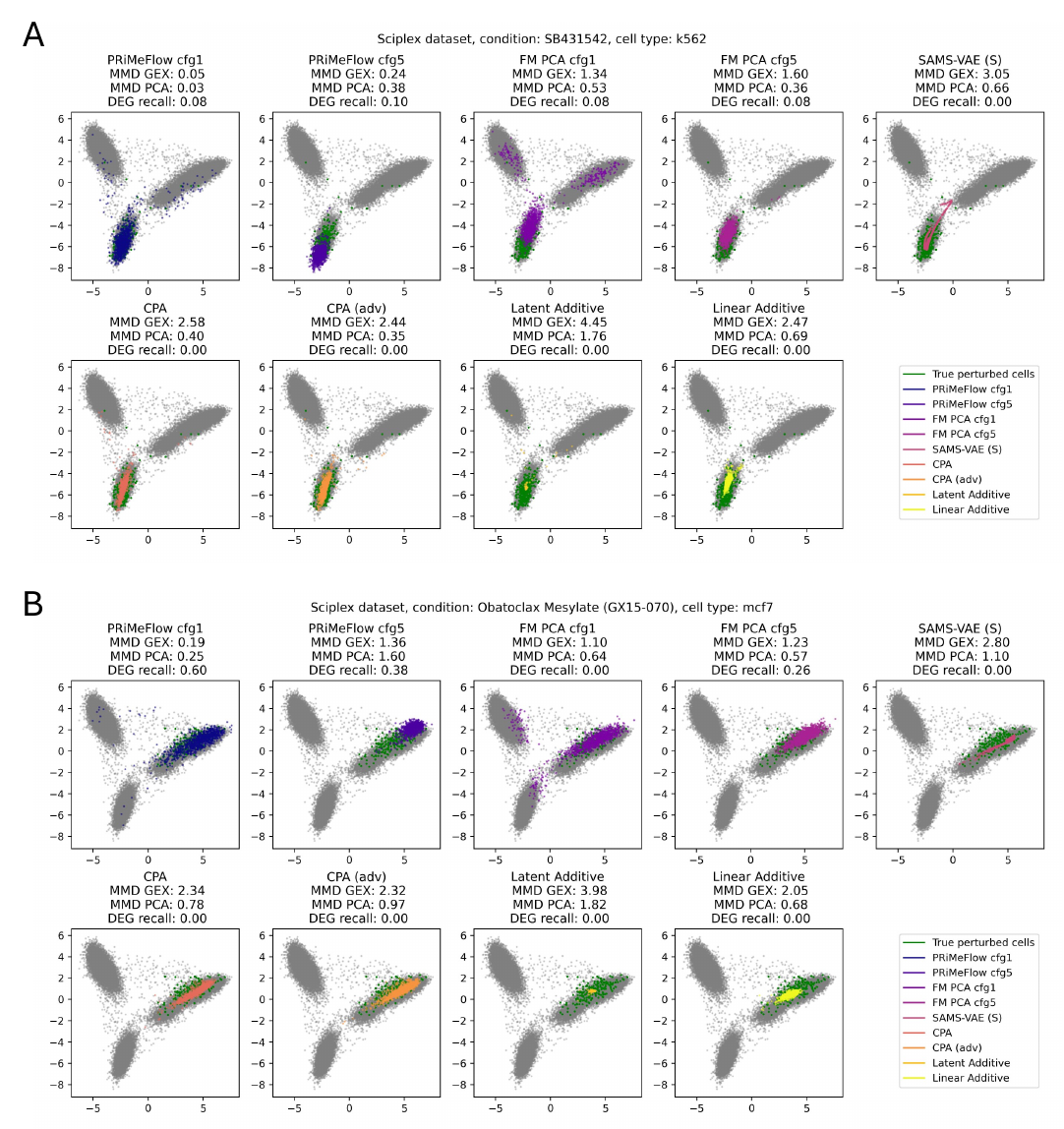}
\caption{Visualization of cells (ground truth and model predictions) in \srivatsan{} test split. Gray marks all the cells in the test split, used for computing PCA and illustrated as background. Green cells are the ground truth perturbed cells for one condition, repeated in each subplot. From each model, 100 cells are sampled, then projected to the same PCA space. \textsf{\bfseries A}): SB431542 perturbation in k562 cell line; \textsf{\bfseries B}) Obatoclax Mesylate (GX15-070) perturbation in mcf7 cell line.}
\label{fig:srivatsan_pca_scatter_2}
\end{figure}

\begin{figure}[t]
\centering
\includegraphics[width=1.\textwidth]{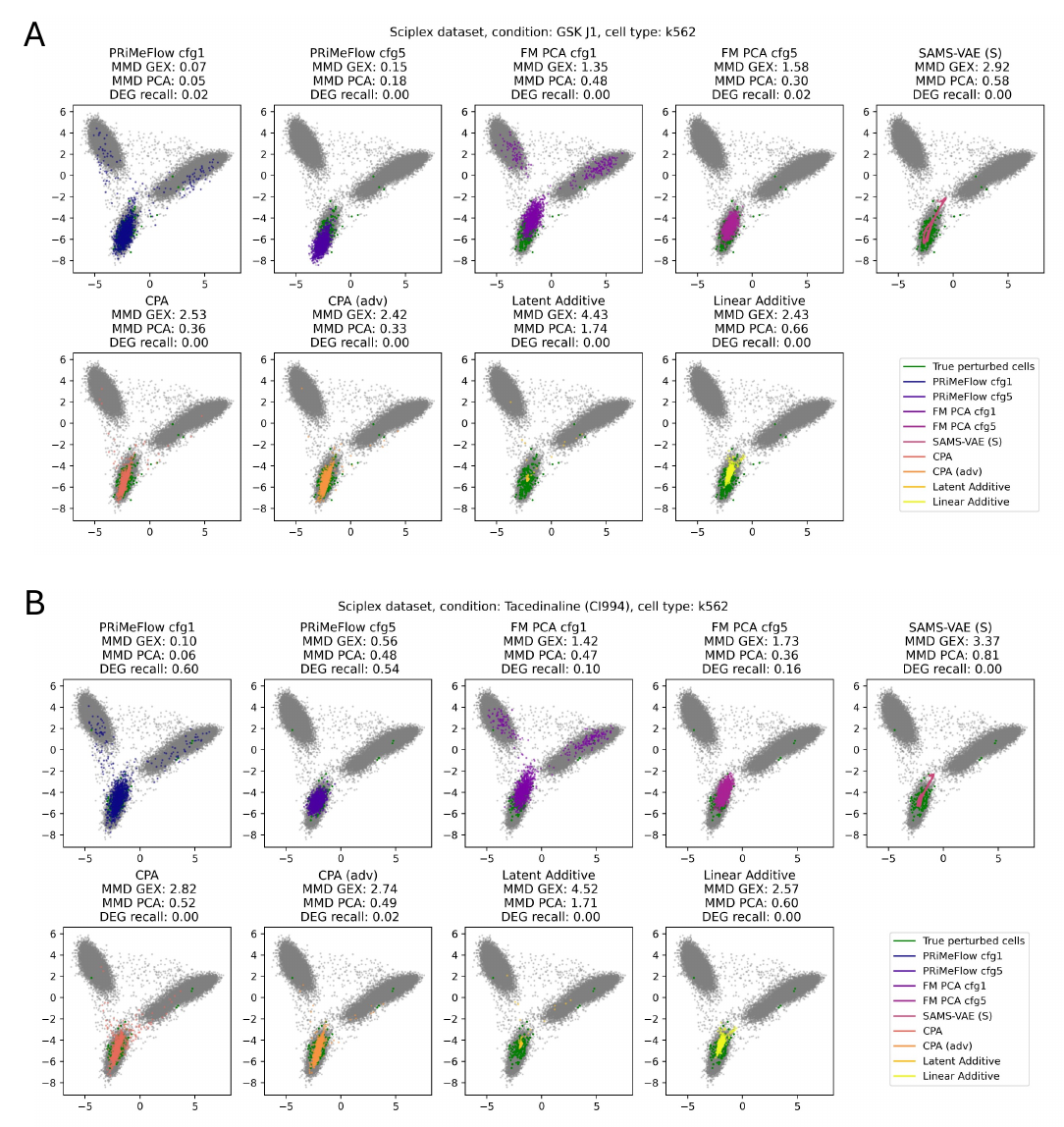}
\caption{Visualization of cells (ground truth and model predictions) in \srivatsan{} test split. Gray marks all the cells in the test split, used for computing PCA and illustrated as background. Green cells are the ground truth perturbed cells for one condition, repeated in each subplot. From each model, 100 cells are sampled, then projected to the same PCA space. \textsf{\bfseries A}): GSK J1 perturbation in k562 cell line; \textsf{\bfseries B}) Tacedinaline (C1994) perturbation in k562 cell line.}
\label{fig:srivatsan_pca_scatter_3}
\end{figure}

\clearpage

\section{PCA visualization for \jiang{} dataset}
\label{sec:add_pca_jiang}

\begin{figure}[!h]
\centering
\includegraphics[width=1.\textwidth]{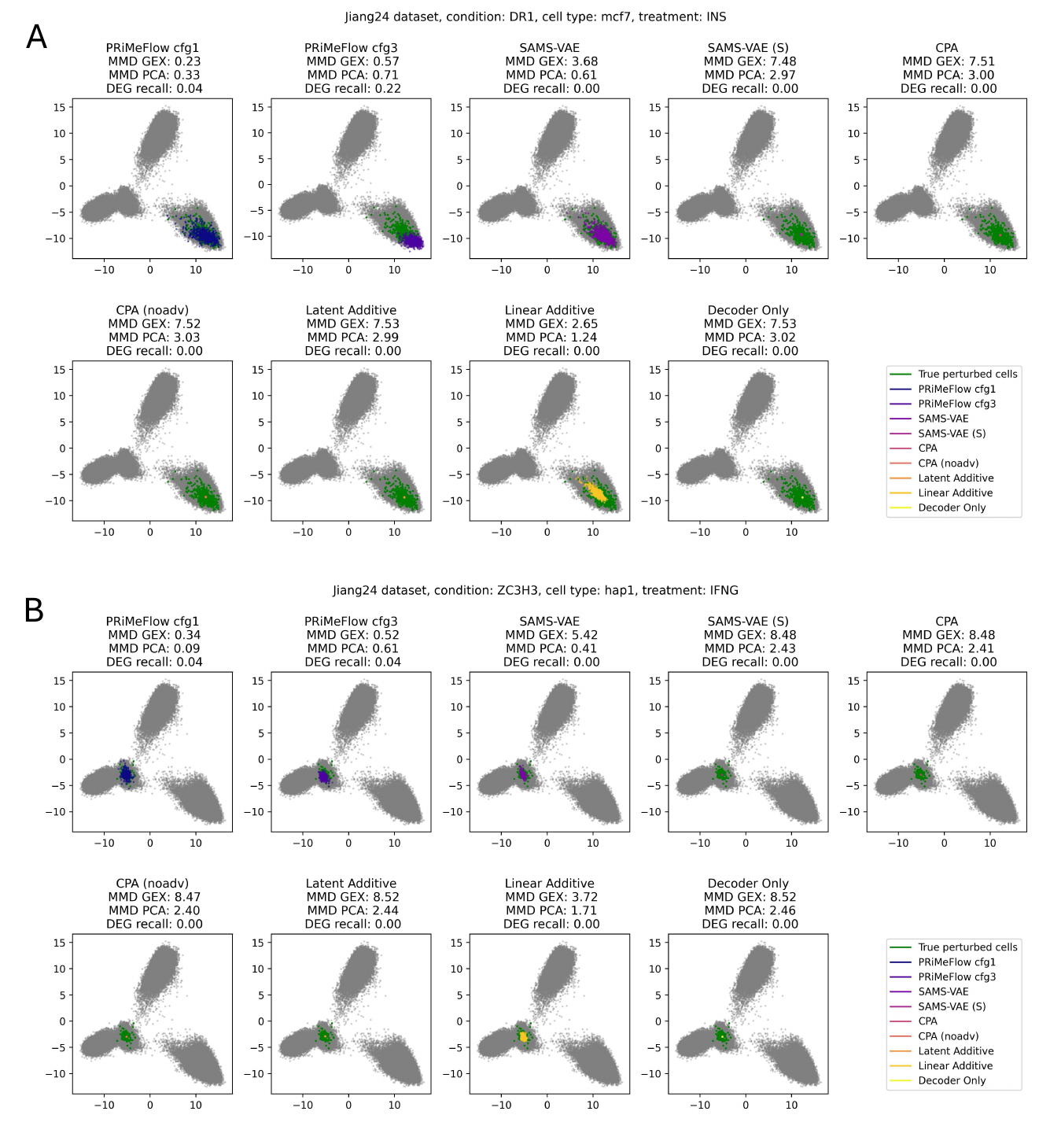}
\caption{Visualization of cells (ground truth and model predictions) in \jiang{} test split. Gray marks all the cells in the test split, used for computing PCA and illustrated as background. Green cells are the ground truth perturbed cells for one condition, repeated in each subplot. From each model, 100 cells are sampled, then projected to the same PCA space. \textsf{\bfseries A}): DR1 perturbation in mcf7 cell line and treated with INS; \textsf{\bfseries B}) ZC3H3 (C1994) perturbation in hap1 cell line and treated with IFNG.}
\label{fig:jiang_pca_scatter}
\end{figure}

\begin{figure}[t]
\centering
\includegraphics[width=1.\textwidth]{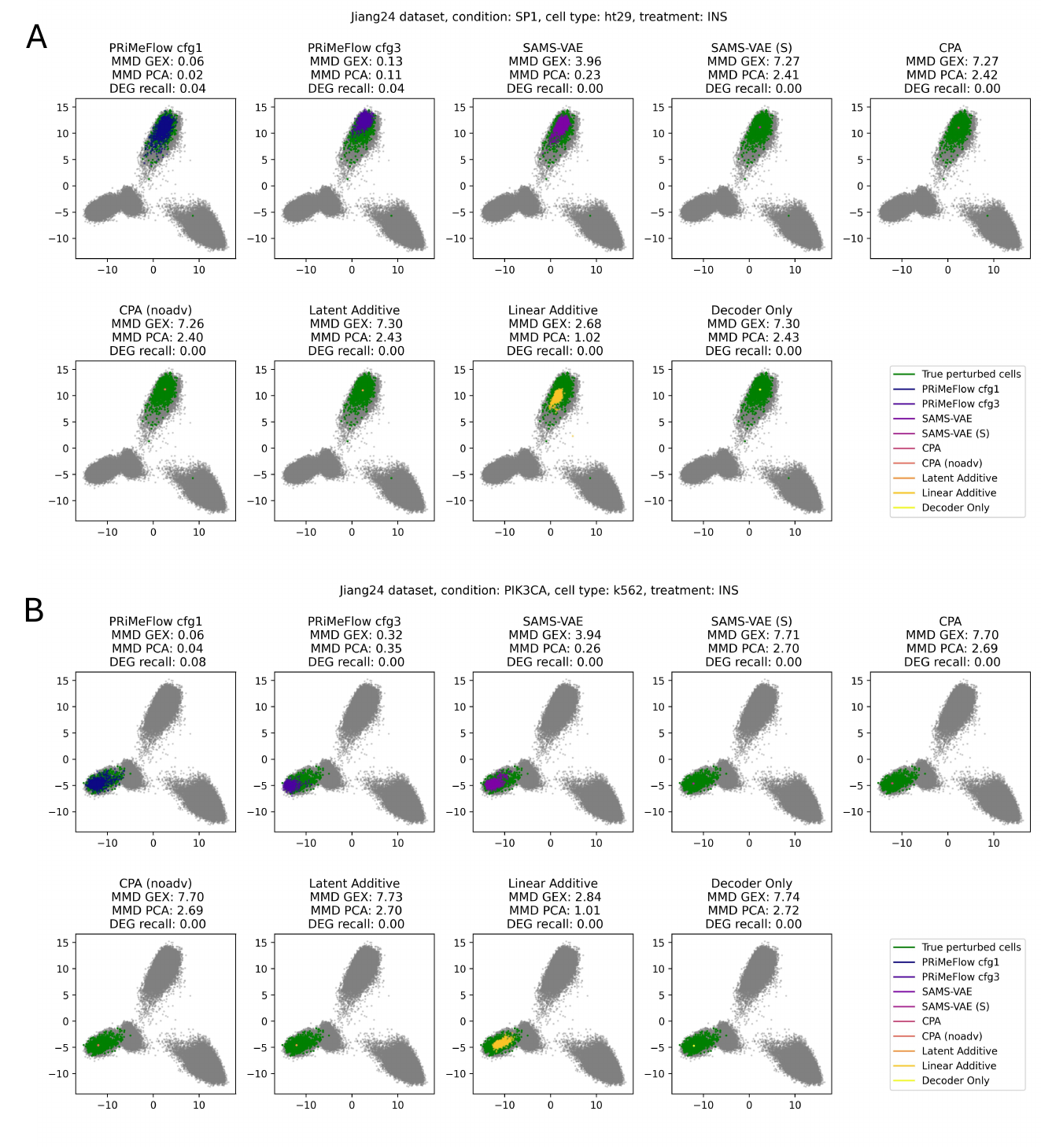}
\caption{Visualization of cells (ground truth and model predictions) in \jiang{} test split. Gray marks all the cells in the test split, used for computing PCA and illustrated as background. Green cells are the ground truth perturbed cells for one condition, repeated in each subplot. From each model, 100 cells are sampled, then projected to the same PCA space. \textsf{\bfseries A}): SP1 perturbation in ht29 cell line and treated with INS; \textsf{\bfseries B}) PIK3CA perturbation in k562 cell line and treated with INS.}
\label{fig:jiang_pca_scatter_2}
\end{figure}

\clearpage

\section{Hyperparameters}
\label{sec:hp}

Hyperparameters for PRiMeFlow MLP and FM PCA are found through 60-trial HPO sweeps with Optuna~\citep{akiba2019optuna}. All activation functions in our models are \texttt{torch.nn.SiLU}~\citep{Paszke2019}.

\begin{table}[h]
\centering
\caption{Manually selected hyperparameters for PRiMeFlow in \srivatsan{} dataset.}
\small
\begin{tabular}{l l l}
\toprule
\textbf{Hyperparameter} & \textbf{Value} & \textbf{Description} \\
\midrule
\texttt{architecture} & U-Net & Architecture choice for the velocity field \\
\texttt{num\_res\_blocks} & 2 & Number of residual blocks after each down/upsampling \\
\texttt{hidden\_dim} & 256 & Base feature dim in U-Net and in input MLPs \\
\texttt{dropout} & 0.1 & Dropout rate \\
\texttt{channel\_mult} & [1, 2, 2, 4, 4] & Hidden dim multipliers at each level \\
\texttt{conv\_resample} & True & Learnable up/downsampling \\
\texttt{num\_heads} & 8 & Number of attention heads \\
\texttt{attention\_resolutions} & [16] & Resolution levels using multihead self-attention \\
\texttt{n\_layers\_gene\_expression} & 2 & MLP layers for gene expression input \\
\texttt{n\_layers\_conditions} & 2 & MLP layers for condition input \\
\texttt{n\_layers\_time} & 2 & MLP layers for time input \\
\texttt{prob\_unconditional} & 0.2 & Probability of dropping out conditions \\
\texttt{interpl\_type} & Linear & Interpolation type in conditional probability path \\
\texttt{sigma} & 0. & Constant noise schedules \\
\texttt{learning\_rate} & 5e-5 & Learning rate \\
\texttt{weight\_decay} & 0. & Weight decay \\
\texttt{batch\_size} & 54 & Batch size\\
\addlinespace
\texttt{num\_gpus} & 4 & Number of GPUs \\
\texttt{grad\_clip\_threshold} & 5.0 & Gradient clipping threshold \\
\texttt{grad\_accum\_batches} & 4 & Number of gradient accumulation steps \\
\bottomrule
\end{tabular}
\end{table}

\begin{table}[h]
\centering
\caption{Manually selected hyperparameters for PRiMeFlow in \norman{} dataset.}
\small
\begin{tabular}{l l l}
\toprule
\textbf{Hyperparameter} & \textbf{Value} & \textbf{Description} \\
\midrule
\texttt{architecture} & U-Net & Architecture choice for the velocity field \\
\texttt{num\_res\_blocks} & 2 & Number of residual blocks after each down/upsampling \\
\texttt{hidden\_dim} & 256 & Base feature dim in U-Net and in input MLPs \\
\texttt{dropout} & 0.1 & Dropout rate \\
\texttt{channel\_mult} & [1, 1, 2, 2, 4] & Hidden dim multipliers at each level \\
\texttt{conv\_resample} & True & Learnable up/downsampling \\
\texttt{num\_heads} & 8 & Number of attention heads \\
\texttt{attention\_resolutions} & [16] & Resolution levels using multihead self-attention \\
\texttt{n\_layers\_gene\_expression} & 2 & MLP layers for gene expression input \\
\texttt{n\_layers\_conditions} & 2 & MLP layers for condition input \\
\texttt{n\_layers\_time} & 2 & MLP layers for time input \\
\texttt{prob\_unconditional} & 0.2 & Probability of dropping out conditions \\
\texttt{interpl\_type} & Linear & Interpolation type in conditional probability path \\
\texttt{sigma} & 0. & Constant noise schedules \\
\texttt{learning\_rate} & 5e-5 & Learning rate \\
\texttt{weight\_decay} & 0. & Weight decay \\
\texttt{batch\_size} & 100 & Batch size\\
\addlinespace
\texttt{num\_gpus} & 4 & Number of GPUs \\
\texttt{grad\_clip\_threshold} & 5.0 & Gradient clipping threshold \\
\texttt{grad\_accum\_batches} & 4 & Number of gradient accumulation steps \\
\bottomrule
\end{tabular}
\end{table}

\begin{table}[h]
\centering
\caption{Manually selected hyperparameters for PRiMeFlow in \jiang{} dataset.}
\small
\begin{tabular}{l l l}
\toprule
\textbf{Hyperparameter} & \textbf{Value} & \textbf{Description} \\
\midrule
\texttt{architecture} & U-Net & Architecture choice for the velocity field \\
\texttt{num\_res\_blocks} & 2 & Number of residual blocks after each down/upsampling \\
\texttt{hidden\_dim} & 256 & Base feature dim in U-Net and in input MLPs \\
\texttt{dropout} & 0.1 & Dropout rate \\
\texttt{channel\_mult} & [1, 1, 2, 2, 4] & Hidden dim multipliers at each level \\
\texttt{conv\_resample} & True & Learnable up/downsampling \\
\texttt{num\_heads} & 8 & Number of attention heads \\
\texttt{attention\_resolutions} & [32] & Resolution levels using multihead self-attention \\
\texttt{n\_layers\_gene\_expression} & 2 & MLP layers for gene expression input \\
\texttt{n\_layers\_conditions} & 2 & MLP layers for condition input \\
\texttt{n\_layers\_time} & 2 & MLP layers for time input \\
\texttt{prob\_unconditional} & 0.2 & Probability of dropping out conditions \\
\texttt{interpl\_type} & Linear & Interpolation type in conditional probability path \\
\texttt{sigma} & 0. & Constant noise schedules \\
\texttt{learning\_rate} & 5e-5 & Learning rate \\
\texttt{weight\_decay} & 0. & Weight decay \\
\texttt{batch\_size} & 38 & Batch size\\
\addlinespace
\texttt{num\_gpus} & 4 & Number of GPUs \\
\texttt{grad\_clip\_threshold} & 5.0 & Gradient clipping threshold \\
\texttt{grad\_accum\_batches} & 4 & Number of gradient accumulation steps \\
\bottomrule
\end{tabular}
\end{table}

\begin{table}[h]
\centering
\caption{HPO selected hyperparameters for PRiMeFlow  MLP in \srivatsan{} dataset.}
\small
\begin{tabular}{l l l}
\toprule
\textbf{Hyperparameter} & \textbf{Value} & \textbf{Description} \\
\midrule
\texttt{architecture} & MLP & Architecture choice for the velocity field \\
\texttt{hidden\_dim} & 3072 & Base feature dim in input MLPs and decoding MLP \\
\texttt{dropout} & 0.0 & Dropout rate \\
\texttt{n\_layers\_gene\_expression} & 1 & MLP layers for gene expression input \\
\texttt{n\_layers\_conditions} & 1 & MLP layers for condition input \\
\texttt{n\_layers\_time} & 3 & MLP layers for time input \\
\texttt{n\_layers\_decoding} & 2 & MLP layers for decoding \\
\texttt{prob\_unconditional} & 0.0 & Probability of dropping out conditions \\
\texttt{interpl\_type} & Linear & Interpolation type in conditional probability path \\
\texttt{sigma} & 0.01 & Constant noise schedules \\
\texttt{learning\_rate} & 3.71e-05 & Learning rate \\
\texttt{weight\_decay} & 4.05e-07 & Weight decay \\
\texttt{batch\_size} & 512 & Batch size\\
\addlinespace
\multicolumn{3}{l}{\textit{Non HPO hyperparameters}} \\
\texttt{num\_gpus} & 1 & Number of GPUs \\
\texttt{grad\_clip\_threshold} & 5.0 & Gradient clipping threshold \\
\texttt{grad\_accum\_batches} & 1 & Number of gradient accumulation steps \\
\bottomrule
\end{tabular}
\end{table}

\begin{table}[h]
\centering
\caption{HPO selected hyperparameters for PRiMeFlow  MLP in \norman{} dataset.}
\small
\begin{tabular}{l l l}
\toprule
\textbf{Hyperparameter} & \textbf{Value} & \textbf{Description} \\
\midrule
\texttt{architecture} & MLP & Architecture choice for the velocity field \\
\texttt{hidden\_dim} & 2048 & Base feature dim in input MLPs and decoding MLP \\
\texttt{dropout} & 0.1 & Dropout rate \\
\texttt{n\_layers\_gene\_expression} & 2 & MLP layers for gene expression input \\
\texttt{n\_layers\_conditions} & 3 & MLP layers for condition input \\
\texttt{n\_layers\_time} & 5 & MLP layers for time input \\
\texttt{n\_layers\_decoding} & 1 & MLP layers for decoding \\
\texttt{prob\_unconditional} & 0.1 & Probability of dropping out conditions \\
\texttt{interpl\_type} & Trigonometric & Interpolation type in conditional probability path \\
\texttt{sigma} & 0.1 & Constant noise schedules \\
\texttt{learning\_rate} & 8.56e-06 & Learning rate \\
\texttt{weight\_decay} & 9.58e-07 & Weight decay \\
\texttt{batch\_size} & 1024 & Batch size\\
\addlinespace
\multicolumn{3}{l}{\textit{Non HPO hyperparameters}} \\
\texttt{num\_gpus} & 1 & Number of GPUs \\
\texttt{grad\_clip\_threshold} & 5.0 & Gradient clipping threshold \\
\texttt{grad\_accum\_batches} & 1 & Number of gradient accumulation steps \\
\bottomrule
\end{tabular}
\end{table}

\begin{table}[h]
\centering
\caption{HPO selected hyperparameters for PRiMeFlow  MLP in \jiang{} dataset.}
\small
\begin{tabular}{l l l}
\toprule
\textbf{Hyperparameter} & \textbf{Value} & \textbf{Description} \\
\midrule
\texttt{architecture} & MLP & Architecture choice for the velocity field \\
\texttt{hidden\_dim} & 3840 & Base feature dim in input MLPs and decoding MLP \\
\texttt{dropout} & 0.1 & Dropout rate \\
\texttt{n\_layers\_gene\_expression} & 2 & MLP layers for gene expression input \\
\texttt{n\_layers\_conditions} & 2 & MLP layers for condition input \\
\texttt{n\_layers\_time} & 2 & MLP layers for time input \\
\texttt{n\_layers\_decoding} & 4 & MLP layers for decoding \\
\texttt{prob\_unconditional} & 0.5 & Probability of dropping out conditions \\
\texttt{interpl\_type} & Linear & Interpolation type in conditional probability path \\
\texttt{sigma} & 0.1 & Constant noise schedules \\
\texttt{learning\_rate} & 2.34e-06 & Learning rate \\
\texttt{weight\_decay} & 3.50e-07 & Weight decay \\
\texttt{batch\_size} & 512 & Batch size\\
\addlinespace
\multicolumn{3}{l}{\textit{Non HPO hyperparameters}} \\
\texttt{num\_gpus} & 1 & Number of GPUs \\
\texttt{grad\_clip\_threshold} & 5.0 & Gradient clipping threshold \\
\texttt{grad\_accum\_batches} & 1 & Number of gradient accumulation steps \\
\bottomrule
\end{tabular}
\end{table}

\begin{table}[h]
\centering
\caption{HPO selected hyperparameters for FM PCA in \srivatsan{} dataset.}
\small
\begin{tabular}{l l l}
\toprule
\textbf{Hyperparameter} & \textbf{Value} & \textbf{Description} \\
\midrule
\texttt{architecture} & MLP & Architecture choice for the velocity field \\
\texttt{hidden\_dim} & 3072 & Base feature dim in input MLPs and decoding MLP \\
\texttt{dropout} & 0.2 & Dropout rate \\
\texttt{n\_layers\_gene\_expression} & 2 & MLP layers for gene expression input \\
\texttt{n\_layers\_conditions} & 3 & MLP layers for condition input \\
\texttt{n\_layers\_time} & 4 & MLP layers for time input \\
\texttt{n\_layers\_decoding} & 4 & MLP layers for decoding \\
\texttt{prob\_unconditional} & 0.4 & Probability of dropping out conditions \\
\texttt{interpl\_type} & Trigonometric & Interpolation type in conditional probability path \\
\texttt{sigma} & 0.1 & Constant noise schedules \\
\texttt{learning\_rate} & 8.18e-05 & Learning rate \\
\texttt{weight\_decay} & 1.28e-10 & Weight decay \\
\texttt{batch\_size} & 1024 & Batch size\\
\addlinespace
\multicolumn{3}{l}{\textit{Non HPO hyperparameters}} \\
\texttt{num\_gpus} & 1 & Number of GPUs \\
\texttt{grad\_clip\_threshold} & 5.0 & Gradient clipping threshold \\
\texttt{grad\_accum\_batches} & 1 & Number of gradient accumulation steps \\
\bottomrule
\end{tabular}
\end{table}

\begin{table}[h]
\centering
\caption{HPO selected hyperparameters for FM PCA in \norman{} dataset.}
\small
\begin{tabular}{l l l}
\toprule
\textbf{Hyperparameter} & \textbf{Value} & \textbf{Description} \\
\midrule
\texttt{architecture} & MLP & Architecture choice for the velocity field \\
\texttt{hidden\_dim} & 1792 & Base feature dim in input MLPs and decoding MLP \\
\texttt{dropout} & 0.2 & Dropout rate \\
\texttt{n\_layers\_gene\_expression} & 4 & MLP layers for gene expression input \\
\texttt{n\_layers\_conditions} & 2 & MLP layers for condition input \\
\texttt{n\_layers\_time} & 2 & MLP layers for time input \\
\texttt{n\_layers\_decoding} & 4 & MLP layers for decoding \\
\texttt{prob\_unconditional} & 0.4 & Probability of dropping out conditions \\
\texttt{interpl\_type} & Trigonometric & Interpolation type in conditional probability path \\
\texttt{sigma} & 0.01 & Constant noise schedules \\
\texttt{learning\_rate} & 1.73e-04 & Learning rate \\
\texttt{weight\_decay} & 6.75e-06 & Weight decay \\
\texttt{batch\_size} & 2048 & Batch size\\
\addlinespace
\multicolumn{3}{l}{\textit{Non HPO hyperparameters}} \\
\texttt{num\_gpus} & 1 & Number of GPUs \\
\texttt{grad\_clip\_threshold} & 5.0 & Gradient clipping threshold \\
\texttt{grad\_accum\_batches} & 1 & Number of gradient accumulation steps \\
\bottomrule
\end{tabular}
\end{table}

\begin{table}[h]
\centering
\caption{HPO selected hyperparameters for FM PCA in \jiang{} dataset.}
\small
\begin{tabular}{l l l}
\toprule
\textbf{Hyperparameter} & \textbf{Value} & \textbf{Description} \\
\midrule
\texttt{architecture} & MLP & Architecture choice for the velocity field \\
\texttt{hidden\_dim} & 3072 & Base feature dim in input MLPs and decoding MLP \\
\texttt{dropout} & 0.0 & Dropout rate \\
\texttt{n\_layers\_gene\_expression} & 5 & MLP layers for gene expression input \\
\texttt{n\_layers\_conditions} & 3 & MLP layers for condition input \\
\texttt{n\_layers\_time} & 4 & MLP layers for time input \\
\texttt{n\_layers\_decoding} & 5 & MLP layers for decoding \\
\texttt{prob\_unconditional} & 0.3 & Probability of dropping out conditions \\
\texttt{interpl\_type} & Linear & Interpolation type in conditional probability path \\
\texttt{sigma} & 0.01 & Constant noise schedules \\
\texttt{learning\_rate} & 8.41e-05 & Learning rate \\
\texttt{weight\_decay} & 1.29e-08 & Weight decay \\
\texttt{batch\_size} & 2048 & Batch size\\
\addlinespace
\multicolumn{3}{l}{\textit{Non HPO hyperparameters}} \\
\texttt{num\_gpus} & 1 & Number of GPUs \\
\texttt{grad\_clip\_threshold} & 5.0 & Gradient clipping threshold \\
\texttt{grad\_accum\_batches} & 1 & Number of gradient accumulation steps \\
\bottomrule
\end{tabular}
\end{table}

\begin{table}[h]
\centering
\caption{Additional hyperparameters for FM PCA OT in \srivatsan{} dataset.}
\small
\begin{tabular}{l l p{6cm}}
\toprule
\texttt{num\_samples\_per\_condition} & 128 & Number of cells sampled per condition to compute the mini-batch OT coupling \\
\texttt{solver\_type} & unbalanced & OT solver algorithm \\
\texttt{tau\_a} & 1.0 & Source marginal constraint weight \\
\texttt{tau\_b} & 1.0 & Target marginal constraint weight \\
\bottomrule
\end{tabular}
\end{table}

\begin{table}[h]
\centering
\caption{Additional hyperparameters for FM PCA OT in \norman{} dataset.}
\small
\begin{tabular}{l l p{6cm}}
\toprule
\texttt{num\_samples\_per\_condition} & 256 & Number of cells sampled per condition to compute the mini-batch OT coupling \\
\texttt{solver\_type} & unbalanced & OT solver algorithm \\
\texttt{tau\_a} & 1.0 & Source marginal constraint weight \\
\texttt{tau\_b} & 1.0 & Target marginal constraint weight \\
\bottomrule
\end{tabular}
\end{table}

\begin{table}[h]
\centering
\caption{Additional hyperparameters for FM PCA OT in \jiang{} dataset.}
\small
\begin{tabular}{l l p{6cm}}
\toprule
\texttt{num\_samples\_per\_condition} & 256 & Number of cells sampled per condition to compute the mini-batch OT coupling \\
\texttt{solver\_type} & unbalanced & OT solver algorithm \\
\texttt{tau\_a} & 1.0 & Source marginal constraint weight \\
\texttt{tau\_b} & 1.0 & Target marginal constraint weight \\
\bottomrule
\end{tabular}
\end{table}

\end{document}